\documentclass[accepted]{uai2025} 
                        
\usepackage[american]{babel}

\usepackage{natbib} 
\usepackage{mathtools} 
\usepackage{booktabs} 
\usepackage{tikz} 
\usepackage{graphicx}
\usepackage{subcaption}
\usepackage{sansmath}
\usepackage{algorithm}
\usepackage{algorithmic}
\usepackage{multirow}
\usepackage{amsfonts}
\DeclareMathOperator*{\concat}{\Vert}

\title{Towards the Next-generation Bayesian Network Classifiers}

\author[1]{Huan Zhang}
\author[2,3]{Daokun Zhang}
\author[1]{Kexin Meng}
\author[3]{Geoffrey I. Webb}
\affil[1]{%
    School of Computer and Artificial Intelligence, Zhengzhou University
}
\affil[2]{%
    School of Computer Science, University of Nottingham Ningbo China
}
\affil[3]{%
    Department of Data Science and AI, Monash University
  }
  
\begin{document}
\onecolumn
\maketitle

\begin{abstract}
Bayesian network classifiers provide a feasible solution to tabular data classification, with a number of merits like high time and memory efficiency, and great explainability. However, due to the parameter explosion and data sparsity issues, Bayesian network classifiers are restricted to low-order feature dependency modeling, making them struggle in extrapolating the occurrence probabilities of complex real-world data. In this paper, we propose a novel paradigm to design high-order Bayesian network classifiers, by learning distributional representations for feature values, as what has been done in word embedding and graph representation learning. The learned distributional representations are encoded with the semantic relatedness between different features through their observed co-occurrence patterns in training data, which then serve as a hallmark to extrapolate the occurrence probabilities of new test samples. As a classifier design realization, we remake the K-dependence Bayesian classifier (KDB) by extending it into a neural version, i.e., NeuralKDB, where a novel neural network architecture is designed to learn distributional representations of feature values and parameterize the conditional probabilities between interdependent features. A stochastic gradient descent based algorithm is designed to train the NeuralKDB model efficiently. Extensive classification experiments on 60 UCI datasets demonstrate that the proposed NeuralKDB classifier excels in capturing high-order feature dependencies and significantly outperforms the conventional Bayesian network classifiers, as well as other competitive classifiers, including two neural network based classifiers without distributional representation learning. 
\end{abstract}

\section{Introduction}
As a typical category of generative classifiers, Bayesian network classifiers (BNCs) \citep{bielza2014discrete} are a powerful tool for tabular data classification, which have the advantages like high training and prediction speed \citep{cheng1999comparing}, support for incremental/continual learning \citep{webb2012learning}, robustness against missing data \citep{sardinha2018revising}, capability in uncertainty quantification with probabilistic predictions \citep{kwon2020uncertainty}, great explainability by uncovering the hidden dependency structures between features \citep{sugahara2024learning}, etc. To learn a BNC, the first step is to construct a directed acyclic graph (DAG) to capture the feature dependency topologies, where nodes represent the random variables corresponding to different features and the class label, while edges describe the dependencies between the random variables. The second step is to estimate a probability distribution for each node conditioned on its parent features and the class label, by checking the feature/label value co-occurrence patterns in training data. 

The complexity of a BNC relies on the assumed dependence order $k$, i.e., how many parent features should each feature depend on at most? Ideally, a BNC with a high dependence order can better capture the hidden feature dependence patterns among the tabular data and yield better prediction performance. However, as the increase of the dependence order $k$, the number of learning parameters of a BNC goes up exponentially. The parameter explosion effect not only restricts the time and memory efficiency of BNCs, but also incurs the data sparsity issue: within the limited amount of training data, it is very difficult to find enough support samples to estimate the high-order conditional probabilities. To alleviate this issue, researchers have to design BNCs with limited dependency orders~\citep{Sahami96, FriedmanGG97, WebbBW05}. 

Like the $n$-gram language models \citep{katz2003estimation}, existing BNCs suffer from the \textbf{curse of dimensionality} \citep{bengio2003neural}: the number of possible feature value combinations in test data is an exponential number, while traditional BNCs can only capture limited dependencies, unable to accurately estimate the joint probability distribution between high-dimensional features and the class label on a support set with a size in an exponential number. To overcome this difficulty, some strategies have been proposed by extending BNCs to capturing high-order feature dependencies, like using an ensemble of several high-order models~\citep{webb2012learning, ZhangPB20} and select the best dependence order according to the support coverage of training data~\citep{MartinezWCZ16}. Though the performance of BNCs can be improved with the adaptations, the existing BNCs are still unable to extrapolate the occurrence probabilities of new test samples that have few overlapped feature value co-occurrence patterns with training data.  

In this paper, we propose a novel paradigm to construct high-order BNCs by learning the distributional representations of feature and label values. The idea of distributional representation learning~\citep{ma2015distributional} is first developed for the task of word embedding~\citep{wang2019evaluating}, based on the ``distributional hypothesis'' in linguistics~\citep{harris1954distributional, firth1957synopsis}: ``{\itshape words that occur in similar contexts tend to have similar meanings}''. Similarly, we can derive the ``distributional hypothesis'' for tabular data: ``{\itshape if two features have similar value co-occurrence patterns with other related features, there exists certain semantic equivalence between the two features}''. For example, the features ``{\itshape high temperature}'' and ``{\itshape hot weather}'' would simultaneously increase the sale of ice creams. As a consequence, we are very likely to frequently observe the feature co-occurrence patterns ``({\itshape high temperature, large ice cream sale})'' and ``({\itshape hot weather, large ice cream sale})'' in the collected data records. It can be anticipated that the ``distributional hypothesis'' will learn similar distributional representations for the features ``{{\itshape high temperature}}'' and ``{\itshape hot weather}''. With the well encoded semantic meanings, the learned feature representations can be used to re-parameterize the probability distribution of each child feature conditioned on its parent features and the class label, which is more flexible and expressive than the traditional BNCs, where each conditional probability value is treated as a separate learnable parameter without any feature semantic linkage. 

Based on the exemplary BNC, K-Dependence Bayesian classifier (KDB)~\citep{Sahami96}, we develop a novel BNC framework--NeuralKDB--by designing a neural network architecture to learn distributional representations of feature and label values and parameterize the conditional probability distributions. We first leverage the $k$-dependence principle in KDB to construct the feature dependence DAG. For each feature, we select its parent features as those that have the top-$k$ mutual information values with it conditioned on the class label. We then design a three-layer neural network to predict the value occurrence probabilities of the child feature conditioned on its parent features and the class label, where the distributional representations of feature and label values are the learnable parameters of the neural network. The neural network is trained with maximum likelihood estimation (MLE), encouraging it to output conditional probability values that are consistent with the feature/label value co-occurrence frequencies in training data. With the well trained neural network, we can obtain the informative distributional representations of feature and label values that well encode the semantic relatedness between them. The occurrence probabilities of test samples can be obtained as the product of the conditional probabilities between interdependent features, which are constructed by the learned distributional representations. The distributional representations can build a correspondence between the non-overlapped feature value co-occurrence patterns across the training and test data by semantic mapping. In this way, the occurrence probabilities of test samples can be accurately extrapolated and their class labels are able to be precisely predicted according to the Bayes rule.  

To train the NeuralKDB model efficiently, we design a parameter optimization algorithm based on stochastic gradient descent. The training algorithm has a time complexity quadratic to the number of features but linear to the number of training samples, which endows the NeuralKDB model with the potential to handle large datasets with a large number of samples and high-dimensional features. Different from the traditional KDB, which suffers from the parameter explosion problem, the number of learnable parameters of NeuralKDB remains unchanged with the increase of the dependence order $k$, which warrants NeuralKDB's high training and prediction efficiency, as well as the robustness to the data sparsity issue. On 60 UCI datasets, we conduct extensive experiments to evaluate the performance of NeuralKDB. The experimental results demonstrate that NeuralKDB excels in capturing high-order feature dependencies, and significantly outperforms the traditional BNCs, the competitive non-Bayesian classifiers like Random Forest, and the neural network based classifiers without distributional representation learning. Our contributions can be summarized as follows

\begin{itemize}
    \item To the best of our knowledge, we are the first to leverage the idea of distributional representation learning to construct high-performance Bayesian network classifiers. 
    \item We develop a novel Bayesian network classifier NeuralKDB by designing a three-layer neural network to learn the distributional representations of feature and label values. 
    \item We conduct extensive experiments on 60 UCI datasets. The results demonstrate that the proposed NeuralKDB significantly outperforms a series of competitive classifiers. 
\end{itemize}

\section{Related Work}

In this section, we review two streams of related work: BNCs and distributional representation learning. 

\subsection{BNCs}
According to the maximum number of allowed parent features, BNCs can be categorized into two groups: one-dependence estimators and K-dependence estimators.

\subsubsection{One-Dependence Estimators}
One-dependence estimators allow each feature to be conditioned on only one parent feature. One-dependence estimators vary in the mechanism used for selecting the parent features. TAN \citep{FriedmanGG97} represents the one-order feature dependencies by constructing a maximum weighted spanning tree, where nodes represent features, and edges represent feature dependencies, whose weights are determined by the mutual information between features conditioned on the class label. Super-Parent TAN \citep{KeoghP99} leverages a wrapper-based method to sequentially construct the tree-structured feature dependencies, by iteratively augmenting the constructed structure with an edge that contributes to the largest classification performance gain. To avoid the computational overhead used for constructing tree dependency structures, AODE \citep{WebbBW05} takes turns to treat each feature as the parent of the remaining features, trains one-dependence estimators with the simple tree structures, and makes predictions using the averaged one-dependence estimators. WAODE \citep{JiangZCW12} augments AODE by assigning different weights to individual one-dependence estimators. 

\subsubsection{K-Dependence Estimators}
The simplicity of one-dependence estimators restricts the performance of BNCs. K-dependence estimators make strides toward capturing high-order feature dependencies. As the pioneered K-dependence estimator, KDB \citep{Sahami96} allows each feature to depend on $k$ parent features that have the largest conditional mutual information with it. Several KDB variants have also been proposed to improve the classification performance. SKDB \citep{MartinezWCZ16} automatically selects the best dependence order for KDB according to the support coverage of training data. MNKDB \citep{KhodayariSamghabadiKT24} constructs a distinct sub-network for each class to accurately capture the intra-class feature dependencies. In addition to the single KDB models, ensemble learning has also been leveraged to design ensembled K-dependence estimators. AnDE \citep{webb2012learning} extends AODE \citep{WebbBW05} to the high-order dependence case, where multiple features jointly serve as the parent of the remaining features when constructing base classifiers. KDF \citep{DuanW17} develops an ensemble variant of KDB, by growing the K-dependence structures in different orders. ESKDB \citep{ZhangPB20} is another ensemble variant of KDB, which imports a stochastic feature discretization method to increase the variance between base models.  

Due to the parameter explosion and data sparsity issues, the existing BNCs are only workable with restricted dependencies, unable to capture the complex feature correlations in real-world data. In this paper, we propose to leverage the idea of distributional representation learning to design the next-generation BNCs, with the expectation to uncover the complete feature dependency structures through feature semantic mapping. 

\subsection{Distributional Representation Learning}
Distributional representation learning has proved its success in the domain of natural language processing (NLP) for the task of word, sentence and document embedding, as well as the domain of graph learning (GL) for the task of graph representation learning. 

\subsubsection{Distributional Representation Learning for NLP}
LSA~\citep{dumais2004latent} and HAL~\citep{lund1996producing} are the first generation of distributional representation learning techniques in NLP. They first establish high-dimensional word representations by constructing the word-document and word-word co-occurrence matrices respectively, then apply dimension reduction techniques like SVD~\citep{lange2010singular} to learn low-dimensional word embeddings. Neural Language Model~\citep{bengio2003neural} leverages a neural network to learn word embeddings for the purpose of probabilistic language modeling, i.e., estimate the probabilities of word sequences. Word2Vec~\citep{mikolov2013distributed} learns word representations by reconstructing context words within a sliding window, through the architectures of CBOW~\citep{mikolov2013efficient} and Skip-Gram~\citep{mikolov2013efficient}. Several Word2Vec variants have also been proposed: GloVe~\citep{pennington2014glove} learns global word-word co-occurrence semantics; FastText~\citep{bojanowski2017enriching} learns sub-word embeddings; Skip-Thought Vectors~\citep{kiros2015skip} and Doc2Vec~\citep{le2014distributed} respectively learn the sentence- and document-level embeddings. Most recently, distributional representation learning has also underpinned the advanced deep lauguage models, like ELMo~\citep{peters2018deep} that learns deep contextual word embeddings from bidirectional LSTMs~\citep{graves2005bidirectional}, and BERT~\citep{devlin2019bert} that leverages marked language modeling to learn contextual word representations with the Transformer architecture~\citep{vaswani2017attention}. 

\subsubsection{Distributional Representation Learning for GL} Based on the ``distributional hypothesis'' on graphs: ``{\itshape if two nodes have overlapped neighborhood structures, they tend to have somewhat structural proximities}''~\citep{zhang2018network}, distributional representation learning has also been employed to learn informative graph node representations. DeepWalk \citep{perozzi2014deepwalk} and Node2Vec~\citep{grover2016node2vec} transform the graph structure into a collection of random walk sequences, and leverage the Skip-Gram model~\citep{mikolov2013efficient} to learn node representations by making an analogy between random walk sequences and natural language sentences. Following the same idea, several random walk based graph representation learning algorithms have been developed. MetaPath2Vec \citep{dong2017metapath2vec} learns node representations for heterogeneous graphs through the meta-path-based random walks. Struct2Vec \citep{ribeiro2017struc2vec} encodes the structural role proximity into node representations. Walklet \citep{perozzi2017don} captures multi-scale structures by skipping nodes in random walks. HARP~\citep{chen2018harp} improves global structure preservation by performing graph coarsening before applying random walks. BiNE~\citep{gao2018bine} uses random walks to model interactions between two node types to learn informative node representations for bipartite graphs. 

Though distributional representation learning has achieved big success for NLP and GL, their application to tabular data learning remains unexplored. In this paper, we import the idea of distributional representation learning into the context of Bayesian classification for the first time, and design a high-performance BNC by unleashing the expressive power of distributional representations.

\section{Problem Definition and Preliminaries}
In this section, we give the formal definition of Bayesian classification and review the preliminaries on the KDB algorithm \citep{Sahami96}. 
\subsection{Problem Definition}
As a general setting for classifier learning, we assume that a set of training samples $\mathcal{T}$ is first given. Each sample $(X,y)\in\mathcal{T}$ is an instantiation of the joint random variable $(\mathsf{X}, \mathsf{y})$, where $\mathsf{X}$ is the $m$-dimensional random feature vector with its $i$th dimension $\mathsf{X}_i$ taking discrete values from the feature value set $\mathcal{A}_i=\{a_{i1},a_{i2},\cdots,a_{i|\mathcal{A}_i|}\}$, while $\mathsf{y}$ is the 1-dimensional random label variable for the class label, taking discrete values from the label value set $\mathcal{Y}=\{y_1,y_2,\cdots,y_{|\mathcal{Y}|}\}$. The objective of Bayesian classification is to estimate the joint probability distribution $\mathbb{P}(\mathsf{X},\mathsf{y})$ from the training samples $\mathcal{T}$, so that we can predict the label of a new test example with feature vector $X^*$ as
\begin{equation}
y^{*}=\mathop{\arg\max}_y\mathbb{P}(\mathsf{X}=X^*,\mathsf{y}=y).
\label{eq:label_pred}
\end{equation}
In general, the training set $\mathcal{T}$ is impossible to cover all possible value instantiations of the joint random variable $(\mathsf{X}, \mathsf{y})$. It is intractable to directly estimation the joint probability distribution $\mathbb{P}(\mathsf{X},\mathsf{y})$ from the training data. Bayesian network classifiers try to decompose the joint probability distribution estimation into a number of conditional probability evaluations among a subset of features and the class label $\mathsf{y}$:
\begin{equation}
\mathbb{P}(\mathsf{X},\mathsf{y})=\mathbb{P}(\mathsf{X}_1,\mathsf{X}_2,\cdots,\mathsf{X}_m,\mathsf{y})=\mathbb{P}(\mathsf{y})\prod_{i=1}^{m}\mathbb{P}(\mathsf{X}_i|\text{Pa}(\mathsf{X}_i),\mathsf{y}),
\label{eq:bayes_classifier}
\end{equation}where $\text{Pa}(\mathsf{X}_i)$ is the set of parent features of $\mathsf{X}_i$, meaning that the random variable $\mathsf{X}_i$ depends on only $\text{Pa}(\mathsf{X}_i)$ and $\mathsf{y}$.  

\subsection{Preliminaries on KDB}
BNCs vary in the construction of Bayesian networks, i.e., how to find the parent features $\text{Pa}(\mathsf{X}_i)$ for each feature $\mathsf{X}_i$. As a representative BNC, KDB selects the parent features $\text{Pa}(\mathsf{X}_i)$ of $\mathsf{X}_i$ as $\mathsf{X}_i$'s most relevant features $\mathsf{X}_j$ that have the top-$k$ conditional mutual information values $\text{MI}(\mathsf{X}_i,\mathsf{X}_j|\mathsf{y})$, which forms the $k$-dependence structure. Algorithm \ref{alg:KDB_network_construct} shows the overall procedure for constructing the $k$-dependence structure. For each feature $\mathsf{X}_i$, its parent feature set $\text{Pa}(\mathsf{X}_i)$ is initialized as the empty set, and its mutual information with $\mathsf{y}$, $\text{MI}(\mathsf{X}_i,\mathsf{y})$, is first calculated from the training set $\mathcal{T}$. The conditional mutual information $\text{MI}(\mathsf{X}_i,\mathsf{X}_j|\mathsf{y})$ between every two distinct features $\mathsf{X}_i$ and $\mathsf{X}_j$ is then computed from  $\mathcal{T}$. After the preparation, the algorithm visits each feature according to the descending order of the mutual information values $\text{MI}(\mathsf{X}_i,\mathsf{y})$. At each visitation, for the currently visited feature $\mathsf{X}_i$, the algorithm selects its parent features $\mathsf{X}_j$ from the previously visited features as those having the top-$k$ conditional mutual information values $\text{MI}(\mathsf{X}_i,\mathsf{X}_j|\mathsf{y})$. At the first $k$ iterations, as the number of previously visited features is smaller than $k$, all of them are selected as the parent features of the visited features, which includes the case that the parent feature set of the first visited feature remains empty. After every feature has been visited, the algorithm completes the selection of parent features, which naturally form a $k$-dependence structure and are finally returned as the algorithm output. 

\begin{algorithm}[h]
\caption{The construction of the $k$-dependence structure}
\label{alg:KDB_network_construct}
\leftline{\textbf{Input}: Training set $\mathcal{T}$ and the dependence order $k$.}
\leftline{\textbf{Output}: $\text{Pa}(\mathsf{X}_i)$ for each $\mathsf{X_i}$.}
\begin{algorithmic}[1] 
\FOR{each feature $\mathsf{X}_i$}
\STATE Initialize its parent feature set $\text{Pa}(\mathsf{X}_i)\leftarrow\emptyset$;
\STATE Compute the mutual information $\text{MI}(\mathsf{X}_i,\mathsf{y})$ with $\mathcal{T}$;
\ENDFOR
\FOR{every two features $\mathsf{X}_i$ and $\mathsf{X}_j$ with $i\neq j$}
\STATE Compute the conditional mutual information $\text{MI}(\mathsf{X}_i,\mathsf{X}_j|\mathsf{y})$ with the training set $\mathcal{T}$;
\ENDFOR
\STATE Initialize the visited feature list $\mathcal{V}\leftarrow\emptyset$; 
\REPEAT
\STATE Select the feature $\mathsf{X}_i\notin\mathcal{V}$ with the largest $\text{MI}(\mathsf{X}_i,\mathsf{y})$ value; 
\STATE Select the top-$\min(k,|\mathcal{V}|)$ features $\mathsf{X}_j$ from $\mathcal{V}$ with the largest $\text{MI}(\mathsf{X}_i,\mathsf{X}_j|\mathsf{y})$ values;
\STATE Update $\mathsf{X}_i$'s parent feature set $\text{Pa}(\mathsf{X}_i)\leftarrow$ the selected top-$\min(k,|\mathcal{V}|)$ features;
\STATE Update the visited feature list $\mathcal{V}\leftarrow\mathcal{V}\cup\{\mathsf{X}_i\}$;
\UNTIL{$\mathcal{V}$ includes all features;}
\RETURN $\text{Pa}(\mathsf{X}_i)$ for each $\mathsf{X}_i$.
\end{algorithmic}
\end{algorithm}

\section{NeuralKDB Classifier}
In this section, we provide the technical details about the proposed NeuralKDB classifier, including how to estimate the conditional probability distribution $\mathbb{P}(\mathsf{X}_i|\text{Pa}(\mathsf{X}_i),\mathsf{y})$ as well as the marginal probability distribution $\mathbb{P}(\mathsf{y})$, how to train a NeuralKDB classifier, and how to make predictions with the trained NeuralKDB classifier.
\subsection{Probability Distribution Estimation}
After the dependence structures between features have been identified, the next step for the construction of BNCs is to estimate the conditional probability distribution $\mathbb{P}(\mathsf{X}_i|\text{Pa}(\mathsf{X}_i),\mathsf{y})$ as well as the marginal probability distribution $\mathbb{P}(\mathsf{y})$. The label probability distribution $\mathbb{P}(\mathsf{y})$ can be easily estimated from the training set $\mathcal{T}$ by counting the proportions of samples belonging to different categories. The conventional KDB classifier estimates the conditional probability $\mathbb{P}(\mathsf{X}_i|\text{Pa}(\mathsf{X}_i),\mathsf{y})$ by observing the value instantiations of the random variable combination $(\mathsf{X}_i,\text{Pa}(\mathsf{X}_i),\mathsf{y})$ from the training set $\mathcal{T}$, which cannot provide a full support for every possible value instantiation of $(\mathsf{X}_i,\text{Pa}(\mathsf{X}_i),\mathsf{y})$ when the dependence order $k$ takes large values, let alone unveil the occurrence probabilities with statistical significance. The inherent defect restricts the conventional KDB classifier to capturing low-order dependencies among features, like $k$ taking the value of 2, limiting its capability in characterizing high-order feature dependencies that broadly occur in the real-world data. 

To overcome this limitation, the proposed NeuralKDB classifier tries to extrapolate the probability $\mathbb{P}(\mathsf{X}_i|\text{Pa}(\mathsf{X}_i),\mathsf{y})$ by learning a vector-formal representation for each feature value and each label value. For each value instantiation of $(\mathsf{X}_i,\text{Pa}(\mathsf{X}_i),\mathsf{y})$, $(x_i,x_{s},y)$ with $\mathsf{X}_i$ and $\mathsf{y}$ taking values $x_i$ and $y$ respectively, each feature $\mathsf{X}_j$ in $\text{Pa}(\mathsf{X}_i)$ taking the corresponding value $x_j\in x_s$, the probability $\mathbb{P}(\mathsf{X}_i=x_i|\text{Pa}(\mathsf{X}_i)=x_s,\mathsf{y}=y)$ is estimated as
\begin{equation}
\mathbb{P}(\mathsf{X}_i=x_i|\text{Pa}(\mathsf{X}_i)=x_s,\mathsf{y}=y)=\frac{\exp(\Phi(x_s,y)\cdot\mathbf{v}^{\prime}_{x_i})}{\sum_{x\in\mathcal{A}_i}\exp(\Phi(x_s,y)\cdot\mathbf{v}^{\prime}_{x})},
\label{eq:condition_prob}
\end{equation}where $\mathbf{v}^{\prime}_{x_i}\in\mathbb{R}^{d}$ is the embedding vector of feature value $x_i$ when its corresponding feature $\mathsf{X}_i$ serves a child feature, while $\Phi(x_s,y)\in\mathbb{R}^{d}$ is the embedding for the value combination $\text{Pa}(\mathsf{X}_i)=x_s$ and $\mathsf{y}=y$, which is constructed by aggregating the embedding vectors of individual feature values $x_j\in x_s$ and the label value $y$:  
\begin{equation}
\Phi(x_s,y)=\big\{\concat_{x_j\in x_s}\mathbf{v}_{x_j}\big\}\concat\mathbf{v}_y,
\end{equation}where $\mathbf{v}_{x_j}\in\mathbb{R}^d$ is the embedding vector of feature value $x_j$ when its corresponding feature $\mathsf{X}_j$ serves as a parent feature and $\mathbf{v}_{y}\in\mathbb{R}^d$ is the embedding vector of label value $y$, while $\concat$ is the aggregation operator implemented by the elementwise sum between two embedding vectors as a common choice. 

\textbf{Neural network based explanation}. The probability modeling mechanism in Eq.(\ref{eq:condition_prob}) is actually achieved through a three-layer neural network. For each parent feature value $x_j\in x_s$, we create a one-hot vector $\mathbf{e}_{x_j}$ in dimension $|\mathcal{A}_j|$, i.e., the number of values of feature $\mathsf{X}_j$, where the $r$th element of $\mathbf{e}_{x_j}$ is 1 if $x_j$ is equal to the $r$th value in $\mathcal{A}_j$, $a_{jr}$, and 0 otherwise. For the label value $y$, we also create a one-hot vector $\mathbf{e}_y$ in dimension $|\mathcal{Y}|$, the number of possible label values, where the $r$th element of $\mathbf{e}_y$ is 1 if $y$ is equal to the $r$th value in $\mathcal{Y}$, $y_{r}$, and 0 otherwise. The one-hot representations for each parent feature value $x_j\in x_s$ and the label value $y$ construct the input layer. 

For each parent feature $\mathsf{X}_j$, we create a weight matrix $\boldsymbol{W}_{\mathsf{X}_j}\in\mathbb{R}^{d\times|\mathcal{A}_j|}$, with its $r$th column being the embedding of the $r$th value in $\mathcal{A}_j$, $\mathbf{v}_{a_{jr}}$, when feature $\mathsf{X}_j$ serves as a parent feature. For label $\mathsf{y}$, we also create a weight matrix $\boldsymbol{W}_\mathsf{y}\in\mathbb{R}^{d\times|\mathcal{Y}|}$, with its $r$th column being the embedding of the $r$th label value in $\mathcal{Y}$, $\mathbf{v}_{y_r}$. With the weight matrices $\boldsymbol{W}_{\mathsf{X}_j}$ and $\boldsymbol{W}_\mathsf{y}$, we have
\begin{equation}
\mathbf{v}_{x_j}=\boldsymbol{W}_{\mathsf{X}_j}\mathbf{e}_{x_j} \;\text{ and }\; \mathbf{v}_y=\boldsymbol{W}_{\mathsf{y}}\mathbf{e}_y.
\end{equation}The embedding vector $\Phi(x_s,y)$ then becomes the hidden-layer representation, which can be reformulated as
\begin{equation}
\Phi(x_s,y)=\big\{\concat_{x_j\in x_s}\boldsymbol{W}_{\mathsf{X}_j}\mathbf{e}_{x_j}\big\}\concat\boldsymbol{W}_{\mathsf{y}}\mathbf{e}_y.
\end{equation}Here, the weight matrices $\boldsymbol{W}_{\mathsf{X}_j}$ for each parent feature $\mathsf{X}_j$ and $\boldsymbol{W}_\mathsf{y}$ for label $\mathsf{y}$ achieve the feature transformation from the input layer to the hidden layer. 

\begin{figure}[t]
\centering
\includegraphics[width=0.475\textwidth]{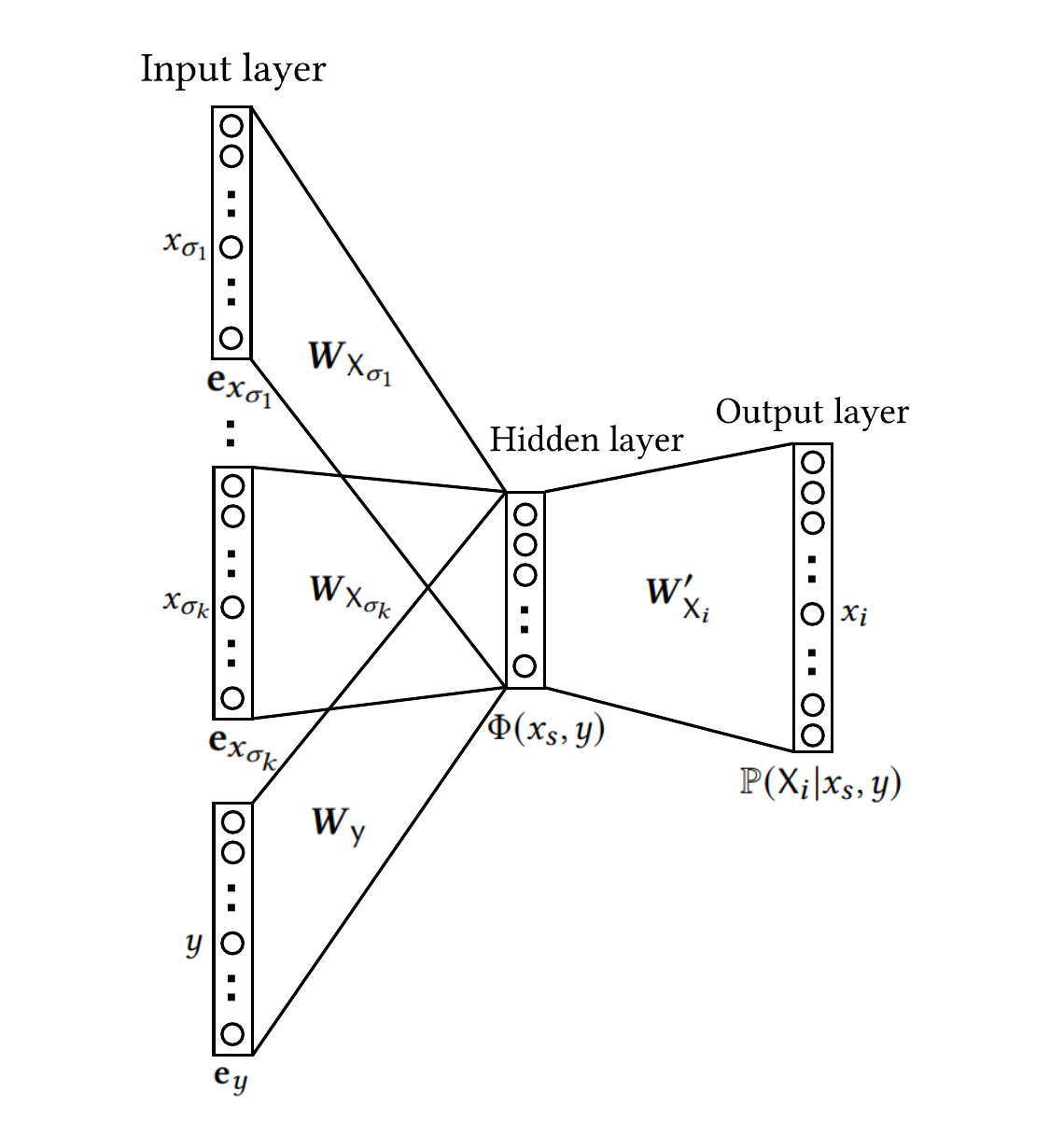}
\caption{The NeuralKDB architecture for modeling the probability $\mathbb{P}(\mathsf{X}_i=x_i|\text{Pa}(\mathsf{X}_i)=x_s,\mathsf{y}=y)$. $\text{Pa}(\mathsf{X}_i)$ is the parent feature set of $\mathsf{X}_i$, with value instantiation $x_s=\{x_{\sigma_1},x_{\sigma_2},\cdots,x_{\sigma_k}\}$. $\mathbf{e}_{x_{\sigma_1}}$, $\mathbf{e}_{x_{\sigma_k}}$ and $\mathbf{e}_{y}$ are the one-hot encodings of the feature/label values $x_{\sigma_1}$, $x_{\sigma_k}$ and $y$ respectively.}
\label{fig:neuralkdb_architecture}
\end{figure}

The probability $\mathbb{P}(\mathsf{X}_i=x_i|\text{Pa}(\mathsf{X}_i)=x_s,\mathsf{y}=y)$ is finally predicted at the output layer, which requires to first transform the hidden-layer representation $\Phi(x_s,y)$ into logits with a linear transformation and then enforce a SoftMax activation. For each child feature $\mathsf{X}_i$, the weight matrix $\boldsymbol{W}_{\mathsf{X}_i}^{\prime}\in\mathbb{R}^{|\mathcal{A}_i|\times d}$ is created to achieve the linear transformation, whose $r$th row is the transpose of the embedding of the $r$th value in $\mathcal{A}_i$, $\mathbf{v}^{\prime}_{a_{ir}}$, when $\mathsf{X}_i$ serves as a child feature. The probability distribution $\mathbb{P}(\mathsf{X}_i|\text{Pa}(\mathsf{X}_i)=x_s,\mathsf{y}=y)$ as a $|\mathcal{A}_i|$-dimensional simplex vector is predicted as 
\begin{equation}
\mathbb{P}(\mathsf{X}_i|\text{Pa}(\mathsf{X}_i)=x_s,\mathsf{y}=y)=\text{SoftMax}\big(\boldsymbol{W}_{\mathsf{X}_i}^{\prime}\Phi(x_s,y)\big),
\end{equation}whose $r$th element is the probability value $\mathbb{P}(\mathsf{X}_i=a_{ir}|\text{Pa}(\mathsf{X}_i)=x_s,\mathsf{y}=y)$ with $\mathsf{X}_i$ taking the value of $a_{ir}$, the $r$th value in $\mathcal{A}_i$. Figure \ref{fig:neuralkdb_architecture} illustrates the neural network architecture,
where the parent feature set $\text{Pa}(\mathsf{X}_i)$ of $\mathsf{X}_i$ is $\{\mathsf{X}_{\sigma_1},\mathsf{X}_{\sigma_2},\cdots,\mathsf{X}_{\sigma_k}\}$, and the corresponding value instantiations are $x_s=\{x_{\sigma_1},x_{\sigma_2},\cdots,x_{\sigma_k}\}$.

\begin{algorithm}[t]
\caption{Training a NeuralKDB classifier}
\label{alg:NeuralKDB_train}
\leftline{\textbf{Input}: Training set $\mathcal{T}$ and the dependence order $k$.}
\leftline{\textbf{Output}: a. The marginal probability distribution $\mathbb{P}(\mathsf{y})$.}
\leftline{b. The parent feature set $\text{Pa}(\mathsf{X}_i)$ for each $\mathsf{X_i}$.}
\leftline{c. The learned embeddings $\{\mathbf{v}^{\prime}_{x},\mathbf{v}_{x},\mathbf{v}_y\}$ for attribute/label values.}
\begin{algorithmic}[1] 
\STATE Estimate the probability distribution $\mathbb{P}(\mathsf{y})$ according to the number of samples belonging to different classes in $\mathcal{T}$;
\STATE Get the parent feature set $\text{Pa}(\mathsf{X}_i)$ for each $\mathsf{X_i}$ by Algorithm \ref{alg:KDB_network_construct};
\STATE Initialize the embeddings $\{\mathbf{v}^{\prime}_{x},\mathbf{v}_{x},\mathbf{v}_y\}\leftarrow$  random numbers;
\STATE Initialize the set of value instantiations $(x_i,x_s,y)$, $\mathcal{S}\leftarrow\emptyset$; 
\FOR{each feature $\mathsf{X}_i$}
\STATE Collect all the value instantiations $(x_i,x_s,y)$ from $\mathcal{T}$ for the random variable combination $(\mathsf{X}_i,\text{Pa}(\mathsf{X}_i),\mathsf{y})$;
\STATE Update $\mathcal{S}\leftarrow\mathcal{S}\;\cup$ collected value instantiations $(x_i,x_s,y)$;
\ENDFOR
\REPEAT
\STATE $\mathcal{B}\leftarrow$ randomly split $\mathcal{S}$ into batches; 
\FOR{each batch in $\mathcal{B}$}
\STATE Compute the partial objective with regard to the value instantiations $(x_i,x_s,y)$ in the batch;
\STATE Update the embeddings $\{\mathbf{v}^{\prime}_{x},\mathbf{v}_{x},\mathbf{v}_y\}\leftarrow$ descend along the gradient of the partial objective; 
\ENDFOR
\UNTIL{a number of epochs expire;}
\RETURN $\mathbb{P}(\mathsf{y})$, $\text{Pa}(\mathsf{X}_i)$ for each $\mathsf{X}_i$, and $\{\mathbf{v}^{\prime}_{x},\mathbf{v}_{x},\mathbf{v}_y\}$.
\end{algorithmic}
\end{algorithm}

\subsection{Model Training}
The parameters $\{\mathbf{v}^{\prime}_{x},\mathbf{v}_{x},\mathbf{v}_y\}$ for each feature value $x$ and each label value $y$ are learned through the principle of maximum likelihood estimation (MLE) from the training set $\mathcal{T}$. By assuming the training samples in $\mathcal{T}$ are independent and identically distributed, the likelihood to be maximized can be formulated as
\begin{equation}
\begin{aligned}
\mathcal{L}&=\prod_{(X,y)\in\mathcal{T}}\mathbb{P}(\mathsf{X}=X,\mathsf{y}=y)\\
&=\prod_{(X,y)\in\mathcal{T}}\Big\{\mathbb{P}(\mathsf{y}=y)\prod_{i=1}^{m}\mathbb{P}(\mathsf{X}_i=x_i|\text{Pa}(\mathsf{X}_i)=x_s,\mathsf{y}=y)\Big\}.
\end{aligned}
\label{eq:likelihood}
\end{equation}
As an equivalence of MLE, we minimize the negative log likelihood to guarantee the numerical stability:
\begin{equation}
\begin{aligned}
-\log \mathcal{L}=&-\sum_{(X,y)\in\mathcal{T}}\sum_{i=1}^{m}\log\mathbb{P}(\mathsf{X}_i=x_i|\text{Pa}(\mathsf{X}_i)=x_s,\mathsf{y}=y)\\
&-\sum_{(X,y)\in\mathcal{T}}\log\mathbb{P}(\mathsf{y}=y).
\label{eq:log_likelihood}
\end{aligned}
\end{equation}
The marginal label probability distribution $\mathbb{P}(\mathsf{y})$ is directly estimated from the training set $\mathcal{T}$ according to the proportions of samples belonging to different classes. This implies that the term $-\sum_{(X,y)\in\mathcal{T}}\log\mathbb{P}(\mathsf{y}=y)$ is a constant with regard to the learnable parameters and can be ignored in the objective function. Finally, we get the learning objective for parameter optimization:
\begin{equation}
\min_{\{\mathbf{v}^{\prime}_{x},\mathbf{v}_{x},\mathbf{v}_y\}}-\sum_{(X,y)\in\mathcal{T}}\sum_{i=1}^{m}\log\mathbb{P}(\mathsf{X}_i=x_i|\text{Pa}(\mathsf{X}_i)=x_s,\mathsf{y}=y).
\label{eq:obj}
\end{equation}

To learn the model parameters effectively, we propose an optimization algorithm based on stochastic gradient descent. Algorithm \ref{alg:NeuralKDB_train} shows the detailed procedure for model training. The marginal probability distribution $\mathbb{P}(\mathsf{y})$ is first estimated from the training set $\mathcal{T}$ by calculating the proportions of samples belonging to different classes. Subsequently, 
the parent feature set $\text{Pa}(\mathsf{X}_i)$ for each feature $\mathsf{X}_i$ is identified from the training set $\mathcal{T}$ using Algorithm \ref{alg:KDB_network_construct}. The embedding vectors $\{\mathbf{v}^{\prime}_{x},\mathbf{v}_{x},\mathbf{v}_y\}$ are then initialized with random numbers. In steps 4-8, we collect the value instantiations $(x_i,x_s,y)$ for every random variable combination $(\mathsf{X}_i,\text{Pa}(\mathsf{X}_i),\mathsf{y})$ from the training set $\mathcal{T}$, and aggregate them into a set $\mathcal{S}$. In steps 9-15, the model parameters $\{\mathbf{v}^{\prime}_{x},\mathbf{v}_{x},\mathbf{v}_y\}$ are updated with stochastic gradient descent, which involves a number of epochs to visit all the value instantiations $(x_i,x_s,y)$ in $\mathcal{S}$. At each epoch, the value instantiations in $\mathcal{S}$ are first randomly split into several batches. The model parameters are then updated by iterating every batch sequentially. At each iteration, the partial objective is first computed as the sum of the negative log probabilities $-\log\mathbb{P}(\mathsf{X}_i=x_i|\text{Pa}(\mathsf{X}_i)=x_s,\mathsf{y}=y)$ for each value instantiation $(x_i,x_s,y)$ in the batch and the embedding vectors $\{\mathbf{v}^{\prime}_{x},\mathbf{v}_{x},\mathbf{v}_y\}$ are then updated by descending along the gradient of the partial objective. 

\textbf{Time complexity analysis}. The time complexity of Algorithm \ref{alg:NeuralKDB_train} is $\mathcal{O}(md|\mathcal{T}|(k+\bar{c})+m^2({\bar{c}}^2|\mathcal{Y}|+k))$, where the number of epochs is taken as a constant, and $\bar{c}=\frac{1}{m}\sum_{i=1}^{m}|\mathcal{A}_i|$ is the average number of feature values. The time complexity is quadratic to the number of features $m$, and the average number of feature values $\bar{c}$, but linear to the number of training samples $|\mathcal{T}|$, 
which demonstrates NeuralKDB's great potential to handle large datasets with a large number of samples and high-dimensional features.

After a NeuralKDB classifier is trained, we can use it to predict the labels of unlabeled test samples. Given an unlabeled sample with feature vector $X^{*}$, NeuralKDB predicts its label as $y^*$ that maximizes the joint probability $\mathbb{P}(\mathsf{X}=X^{*},\mathsf{y}=y^{*})$ as in Eq.(\ref{eq:label_pred}).

\section{Experiments}
\label{sec: Experiments and Results}
In this section, we carry out a series of experiments to evaluate the performance of the proposed NeuralKDB classifier.
\subsection{Experimental Settings}
\label{sec: Experimental setting and benchmark data}
\paragraph{Datasets.} To make a comprehensive evaluation, we collect 60 public tabular datasets from the UCI repository \citep{asuncion2007uci}, followed by some pre-processing operations, including filling missing values, discretizing numeric features, and removing redundant features.

\paragraph{Baselines.} We compare NeuralKDB with a series of baselines:

\begin{itemize}
    \item \textbf{KDF} \citep{DuanW17} is a K-dependence estimator ensemble, where the dependence structure of each base estimator is constructed with a different feature selection order. 
    \item \textbf{KDB} \citep{Sahami96} is the counterpart of NeuralKDB, which simply uses the normalized feature-feature co-occurrence frequencies as the conditional probability estimates.
    \item \textbf{TAN} \citep{FriedmanGG97} is an one-dependence estimator, whose dependence structure is a maximum weighted spanning tree. 
    \item \textbf{Naive Bayes (NB)} \citep{Webb10} is a simple yet effective BNC constructed by the naive feature independence assumption. 
    \item \textbf{Random Forest (RF)} \citep{Breiman01} is a forest-structured classifier, as an ensemble of a number of decision trees. 
    \item \textbf{NeuralNB} is a degraded version of NeuralKDB, where child features depend on only the class label. 
    \item \textbf{Neural Network (NN)} directly predicts class labels from the latent representations of samples, which are obtained by the concatenation of feature value embeddings. 
\end{itemize}
Among the baselines, KDF, KDB, TAN and NB are the conventional BNCs, and RF is a powerful classifier that is widely used in many real-world applications. NeuralNB and NN are the neural network based classifiers without distributional representation learning. 

\paragraph{Implementation Details.} The embedding dimension, batch size and epoch number for the implementations of NeuralKDB, NeuralNB and NN are consistently set to 128, 32, and 10 respectively. The number of dependence order $k$ for the K-dependence estimators (NeuralKDB, KDF and KDB) is consistently set to 2.

\begin{table}[!t]
\caption{\label{tab:AccuracyComparison}Classification accuracy comparison between NeuralKDB and baseline classifiers on 60 UCI datasets.}
\scriptsize
{\centering \begin{tabular*}{\textwidth}{lr@{\hspace{0cm}}c@{\hspace{0cm}}rr@{\hspace{0cm}}c@{\hspace{0cm}}r@{\hspace{0.0cm}}cr@{\hspace{0cm}}c@{\hspace{0cm}}r@{\hspace{0.0cm}}cr@{\hspace{0cm}}c@{\hspace{0cm}}r@{\hspace{0.0cm}}cr@{\hspace{0cm}}c@{\hspace{0cm}}r@{\hspace{0.0cm}}cr@{\hspace{0cm}}c@{\hspace{0cm}}r@{\hspace{0.0cm}}cr@{\hspace{0cm}}c@{\hspace{0cm}}r@{\hspace{0.0cm}}cr@{\hspace{0cm}}c@{\hspace{0cm}}r@{\hspace{0.0cm}}c}
\hline
Dataset & \multicolumn{3}{c}{NeuralKDB}& \multicolumn{4}{c}{NeuralNB} & \multicolumn{4}{c}{NN} & \multicolumn{4}{c}{KDF} & \multicolumn{4}{c}{KDB} & \multicolumn{4}{c}{TAN} & \multicolumn{4}{c}{NB} & \multicolumn{4}{c}{RF} \\
\hline
anneal.ORIG&\textbf{92.71}&$\pm$&0.97&89.37&$\pm$&1.16&$\bullet$&78.74&$\pm$&2.92&$\bullet$&90.71&$\pm$&1.31&$\bullet$&\underline{92.34}&$\pm$&0.42&&92.27&$\pm$&1.32&&88.55&$\pm$&1.47&$\bullet$&91.38&$\pm$&1.77&\\
anneal&\textbf{99.03}&$\pm$&0.20&94.42&$\pm$&1.78&$\bullet$&90.56&$\pm$&2.52&$\bullet$&98.22&$\pm$&0.72&$\bullet$&\underline{98.36}&$\pm$&0.56&$\bullet$&98.29&$\pm$&0.50&$\bullet$&96.13&$\pm$&0.56&$\bullet$&98.29&$\pm$&0.56&$\bullet$\\
artificial-characters&59.92&$\pm$&1.03&34.80&$\pm$&1.13&$\bullet$&59.93&$\pm$&1.60&&58.86&$\pm$&0.55&$\bullet$&\underline{62.60}&$\pm$&0.43&$\circ$&57.83&$\pm$&0.65&$\bullet$&35.88&$\pm$&0.95&$\bullet$&\textbf{67.69}&$\pm$&0.78&$\circ$\\
audiology&\textbf{81.47}&$\pm$&6.29&\underline{76.47}&$\pm$&7.78&&23.82&$\pm$&6.19&$\bullet$&68.82&$\pm$&2.83&$\bullet$&64.41&$\pm$&6.36&$\bullet$&68.82&$\pm$&2.18&$\bullet$&75.59&$\pm$&6.94&$\bullet$&69.71&$\pm$&5.75&\\
autos&68.20&$\pm$&8.25&53.11&$\pm$&6.92&$\bullet$&49.51&$\pm$&5.36&$\bullet$&67.87&$\pm$&6.52&&\underline{69.51}&$\pm$&5.51&&67.54&$\pm$&6.70&&53.44&$\pm$&7.48&$\bullet$&\textbf{70.82}&$\pm$&6.60&\\
balance-scale&85.13&$\pm$&3.08&87.49&$\pm$&2.64&&\underline{89.20}&$\pm$&1.83&$\circ$&85.99&$\pm$&2.25&&73.05&$\pm$&3.06&$\bullet$&83.53&$\pm$&2.22&&\textbf{89.41}&$\pm$&1.03&$\circ$&75.40&$\pm$&3.23&$\bullet$\\
breast-cancer&70.00&$\pm$&1.52&70.93&$\pm$&2.33&&\textbf{73.49}&$\pm$&2.24&$\circ$&69.53&$\pm$&5.03&&66.05&$\pm$&3.98&$\bullet$&68.60&$\pm$&4.11&&\underline{72.79}&$\pm$&1.04&$\circ$&67.67&$\pm$&3.23&\\
breast-w&96.57&$\pm$&0.62&\textbf{97.52}&$\pm$&0.40&$\circ$&95.90&$\pm$&1.29&&96.29&$\pm$&1.03&&95.05&$\pm$&0.54&$\bullet$&96.95&$\pm$&0.43&&\underline{97.24}&$\pm$&0.62&$\circ$&95.62&$\pm$&1.63&\\
car&92.55&$\pm$&1.20&83.94&$\pm$&1.74&$\bullet$&\textbf{95.25}&$\pm$&0.56&$\circ$&93.82&$\pm$&0.60&&\underline{94.94}&$\pm$&1.31&$\circ$&93.82&$\pm$&0.27&&85.10&$\pm$&1.08&$\bullet$&91.39&$\pm$&1.02&\\
cardiotocography&90.22&$\pm$&0.88&83.54&$\pm$&1.04&$\bullet$&90.44&$\pm$&1.11&&90.16&$\pm$&0.90&&\underline{90.69}&$\pm$&0.71&&88.97&$\pm$&0.81&$\bullet$&83.89&$\pm$&1.37&$\bullet$&\textbf{90.94}&$\pm$&0.76&$\circ$\\
climate&85.68&$\pm$&2.81&86.91&$\pm$&2.16&&\textbf{90.49}&$\pm$&0.70&$\circ$&89.63&$\pm$&1.66&$\circ$&75.31&$\pm$&6.36&$\bullet$&88.40&$\pm$&1.66&&87.90&$\pm$&0.94&&\underline{90.12}&$\pm$&0.44&$\circ$\\
colic.ORIG&72.55&$\pm$&3.72&74.00&$\pm$&4.48&&66.55&$\pm$&3.72&$\bullet$&\textbf{74.91}&$\pm$&1.52&&\textbf{74.91}&$\pm$&1.04&&74.73&$\pm$&1.19&&74.55&$\pm$&1.44&&68.91&$\pm$&1.63&\\
colic&\underline{80.00}&$\pm$&2.49&77.27&$\pm$&2.32&$\bullet$&79.45&$\pm$&1.38&&79.09&$\pm$&2.87&&73.82&$\pm$&4.33&$\bullet$&78.73&$\pm$&2.09&&78.91&$\pm$&1.97&&\textbf{80.55}&$\pm$&3.38&\\
connectionist-vowel&\underline{80.00}&$\pm$&2.99&71.77&$\pm$&3.87&$\bullet$&60.25&$\pm$&4.26&$\bullet$&79.62&$\pm$&3.73&&78.35&$\pm$&5.05&&78.61&$\pm$&4.14&&74.05&$\pm$&2.72&$\bullet$&\textbf{81.90}&$\pm$&3.76&\\
credit-a&\textbf{86.47}&$\pm$&3.13&84.54&$\pm$&3.11&&\underline{86.09}&$\pm$&3.08&&82.80&$\pm$&3.62&$\bullet$&77.78&$\pm$&3.02&$\bullet$&80.87&$\pm$&3.83&$\bullet$&\underline{86.09}&$\pm$&2.93&&84.73&$\pm$&2.97&\\
credit-g&74.40&$\pm$&2.10&73.80&$\pm$&1.71&&73.87&$\pm$&3.02&&\textbf{75.40}&$\pm$&2.47&&72.27&$\pm$&1.34&$\bullet$&73.27&$\pm$&4.04&&\underline{74.67}&$\pm$&1.94&&71.93&$\pm$&3.46&\\
cylinder-bands&69.63&$\pm$&5.81&\underline{78.27}&$\pm$&1.71&&65.56&$\pm$&2.11&&76.05&$\pm$&3.79&&77.28&$\pm$&2.33&$\circ$&75.93&$\pm$&2.76&$\circ$&\textbf{78.77}&$\pm$&2.45&$\circ$&69.63&$\pm$&1.71&\\
diabetes&\underline{75.39}&$\pm$&2.33&75.30&$\pm$&2.29&&75.04&$\pm$&2.47&&74.43&$\pm$&3.07&&68.87&$\pm$&3.38&$\bullet$&75.13&$\pm$&2.97&&\textbf{77.30}&$\pm$&2.25&&73.65&$\pm$&4.84&\\
ecoli&81.39&$\pm$&4.97&\underline{81.98}&$\pm$&6.84&&72.87&$\pm$&6.29&$\bullet$&75.45&$\pm$&9.43&&69.50&$\pm$&7.46&$\bullet$&75.45&$\pm$&7.42&$\bullet$&\textbf{82.97}&$\pm$&3.99&&73.66&$\pm$&6.85&$\bullet$\\
energy-y1&61.91&$\pm$&2.41&44.70&$\pm$&0.19&$\bullet$&56.43&$\pm$&3.22&$\bullet$&61.91&$\pm$&2.93&&\textbf{66.52}&$\pm$&1.37&$\circ$&\underline{65.74}&$\pm$&1.35&$\circ$&49.91&$\pm$&2.41&$\bullet$&63.74&$\pm$&2.47&\\
energy-y2&\textbf{51.91}&$\pm$&1.59&47.30&$\pm$&2.03&$\bullet$&\underline{51.04}&$\pm$&3.51&&50.78&$\pm$&2.49&&48.61&$\pm$&2.43&$\bullet$&49.30&$\pm$&2.25&&48.52&$\pm$&2.96&$\bullet$&\underline{51.04}&$\pm$&3.92&\\
glass&\underline{57.19}&$\pm$&8.60&53.75&$\pm$&5.91&&45.31&$\pm$&8.77&&\underline{57.19}&$\pm$&3.24&&\textbf{58.44}&$\pm$&7.04&&55.94&$\pm$&5.99&&55.00&$\pm$&6.94&&56.25&$\pm$&6.72&\\
hayes-roth&75.00&$\pm$&5.31&\underline{80.00}&$\pm$&8.28&&69.17&$\pm$&3.73&&71.67&$\pm$&5.43&&68.75&$\pm$&5.51&&70.42&$\pm$&3.42&&\textbf{82.50}&$\pm$&7.00&$\circ$&75.00&$\pm$&4.66&\\
heart-c&\underline{82.64}&$\pm$&1.20&80.88&$\pm$&4.70&&80.22&$\pm$&2.06&$\bullet$&77.36&$\pm$&2.14&$\bullet$&68.79&$\pm$&4.30&$\bullet$&73.85&$\pm$&2.95&$\bullet$&\textbf{83.74}&$\pm$&4.35&&74.29&$\pm$&2.14&$\bullet$\\
heart-h&82.05&$\pm$&3.63&\textbf{82.95}&$\pm$&4.25&&\underline{82.27}&$\pm$&4.93&&77.05&$\pm$&2.19&$\bullet$&75.00&$\pm$&2.13&$\bullet$&76.82&$\pm$&2.22&$\bullet$&\underline{82.27}&$\pm$&3.27&&80.68&$\pm$&5.08&\\
heart-statlog&80.99&$\pm$&3.66&\underline{83.95}&$\pm$&3.02&&81.48&$\pm$&3.02&&78.27&$\pm$&4.42&&72.35&$\pm$&5.63&$\bullet$&77.78&$\pm$&3.81&&\textbf{85.19}&$\pm$&2.47&&76.30&$\pm$&4.31&\\
hepatitis&\underline{87.39}&$\pm$&4.71&\underline{87.39}&$\pm$&5.41&&80.00&$\pm$&5.41&&86.09&$\pm$&5.00&&85.65&$\pm$&4.51&&84.35&$\pm$&2.83&&\textbf{89.13}&$\pm$&3.07&&82.61&$\pm$&4.61&\\
hypothyroid&\underline{93.07}&$\pm$&0.56&92.65&$\pm$&1.21&&\textbf{93.53}&$\pm$&0.83&&92.30&$\pm$&0.53&$\bullet$&92.05&$\pm$&0.66&$\bullet$&92.28&$\pm$&0.57&$\bullet$&92.84&$\pm$&0.62&$\bullet$&92.51&$\pm$&0.88&\\
ionosphere&\underline{92.57}&$\pm$&1.83&90.67&$\pm$&2.73&&86.67&$\pm$&3.37&$\bullet$&92.19&$\pm$&1.70&&\textbf{92.76}&$\pm$&1.09&&91.24&$\pm$&2.47&&91.43&$\pm$&3.63&&90.86&$\pm$&1.59&\\
iris&\textbf{95.11}&$\pm$&3.30&92.00&$\pm$&3.72&&92.89&$\pm$&5.75&&93.78&$\pm$&3.98&&\underline{94.22}&$\pm$&3.72&&92.44&$\pm$&4.61&&91.11&$\pm$&4.71&&93.33&$\pm$&6.09&\\
kr-vs-kp&95.58&$\pm$&0.69&85.99&$\pm$&0.74&$\bullet$&\underline{96.89}&$\pm$&1.25&$\circ$&94.56&$\pm$&1.23&&95.83&$\pm$&0.80&&92.26&$\pm$&0.92&$\bullet$&88.43&$\pm$&0.86&$\bullet$&\textbf{98.75}&$\pm$&0.32&$\circ$\\
labor&\textbf{91.76}&$\pm$&6.71&\underline{90.59}&$\pm$&5.26&&70.59&$\pm$&11.0&$\bullet$&88.24&$\pm$&5.88&&82.35&$\pm$&5.88&$\bullet$&88.24&$\pm$&7.20&&89.41&$\pm$&7.67&&84.71&$\pm$&13.5&\\
letter&85.08&$\pm$&0.41&68.40&$\pm$&0.39&$\bullet$&82.70&$\pm$&0.93&$\bullet$&84.62&$\pm$&0.47&&\underline{87.48}&$\pm$&0.27&$\circ$&83.11&$\pm$&0.37&$\bullet$&70.32&$\pm$&0.25&$\bullet$&\textbf{87.54}&$\pm$&0.53&$\circ$\\
libras&59.81&$\pm$&1.06&64.63&$\pm$&3.67&$\circ$&24.63&$\pm$&4.12&$\bullet$&\underline{71.11}&$\pm$&4.69&$\circ$&\textbf{71.30}&$\pm$&3.82&$\circ$&70.19&$\pm$&3.90&$\circ$&63.70&$\pm$&3.61&&57.04&$\pm$&2.50&\\
lymph&\textbf{87.73}&$\pm$&4.13&\underline{83.64}&$\pm$&5.43&&75.45&$\pm$&3.37&$\bullet$&82.73&$\pm$&3.45&&81.82&$\pm$&5.33&$\bullet$&81.36&$\pm$&5.43&$\bullet$&82.73&$\pm$&4.71&&78.64&$\pm$&6.14&$\bullet$\\
mfeat-f&\textbf{81.23}&$\pm$&1.83&76.50&$\pm$&0.79&$\bullet$&60.67&$\pm$&2.60&$\bullet$&77.57&$\pm$&1.73&$\bullet$&71.97&$\pm$&1.81&$\bullet$&\underline{78.40}&$\pm$&1.65&$\bullet$&77.30&$\pm$&1.40&$\bullet$&69.37&$\pm$&3.42&$\bullet$\\
monks&97.01&$\pm$&2.28&70.90&$\pm$&2.18&$\bullet$&87.66&$\pm$&4.20&$\bullet$&\textbf{100.00}&$\pm$&0.00&$\circ$&99.64&$\pm$&0.80&&\textbf{100.00}&$\pm$&0.00&$\circ$&75.21&$\pm$&2.34&$\bullet$&97.72&$\pm$&1.66&\\
mushroom&99.98&$\pm$&0.02&96.41&$\pm$&0.59&$\bullet$&\textbf{100.00}&$\pm$&0.00&&\textbf{100.00}&$\pm$&0.00&&\textbf{100.00}&$\pm$&0.00&&\textbf{100.00}&$\pm$&0.00&&97.50&$\pm$&0.31&$\bullet$&\textbf{100.00}&$\pm$&0.00&\\
newthyroid&\underline{93.13}&$\pm$&1.78&\textbf{93.75}&$\pm$&1.56&&85.00&$\pm$&1.40&$\bullet$&92.50&$\pm$&1.31&&92.50&$\pm$&2.04&&89.06&$\pm$&1.10&$\bullet$&91.88&$\pm$&1.71&&91.88&$\pm$&2.04&\\
optdigits&\textbf{96.06}&$\pm$&0.60&92.08&$\pm$&0.47&$\bullet$&90.90&$\pm$&0.71&$\bullet$&94.50&$\pm$&0.56&$\bullet$&93.02&$\pm$&0.98&$\bullet$&\underline{94.82}&$\pm$&0.56&$\bullet$&92.42&$\pm$&0.51&$\bullet$&85.73&$\pm$&0.83&$\bullet$\\
page-blocks&\textbf{94.03}&$\pm$&0.54&91.23&$\pm$&0.90&$\bullet$&93.68&$\pm$&0.51&$\bullet$&93.50&$\pm$&0.27&$\bullet$&\underline{93.86}&$\pm$&0.30&&92.84&$\pm$&0.39&$\bullet$&92.97&$\pm$&0.37&$\bullet$&93.30&$\pm$&0.40&\\
parkinsons&\textbf{90.34}&$\pm$&5.80&74.48&$\pm$&6.38&$\bullet$&80.69&$\pm$&2.56&$\bullet$&78.62&$\pm$&5.67&$\bullet$&78.28&$\pm$&5.80&$\bullet$&77.59&$\pm$&5.03&$\bullet$&76.90&$\pm$&5.80&$\bullet$&\underline{84.83}&$\pm$&2.56&\\
pendigits&\underline{97.45}&$\pm$&0.16&86.72&$\pm$&0.82&$\bullet$&95.49&$\pm$&1.36&$\bullet$&96.71&$\pm$&0.24&$\bullet$&\textbf{97.64}&$\pm$&0.07&$\circ$&96.64&$\pm$&0.12&$\bullet$&88.13&$\pm$&0.53&$\bullet$&96.27&$\pm$&0.25&$\bullet$\\
primary-tumor&\textbf{46.27}&$\pm$&5.43&42.35&$\pm$&4.57&$\bullet$&35.69&$\pm$&2.82&$\bullet$&43.33&$\pm$&5.02&$\bullet$&41.76&$\pm$&7.29&$\bullet$&43.53&$\pm$&5.48&&\underline{44.31}&$\pm$&6.70&&38.04&$\pm$&4.57&$\bullet$\\
qar-biodegradation&82.85&$\pm$&1.29&79.87&$\pm$&1.49&$\bullet$&82.66&$\pm$&2.81&&\underline{83.04}&$\pm$&2.27&&\textbf{83.61}&$\pm$&1.31&&82.03&$\pm$&2.37&&79.37&$\pm$&1.67&$\bullet$&82.59&$\pm$&1.43&\\
robot-24&89.88&$\pm$&0.91&79.76&$\pm$&0.43&$\bullet$&\underline{92.55}&$\pm$&1.31&$\circ$&87.65&$\pm$&0.80&$\bullet$&90.23&$\pm$&0.71&&88.05&$\pm$&0.79&$\bullet$&80.39&$\pm$&0.83&$\bullet$&\textbf{93.83}&$\pm$&0.70&$\circ$\\
segment&95.18&$\pm$&0.96&89.26&$\pm$&0.52&$\bullet$&93.54&$\pm$&1.68&&\underline{95.47}&$\pm$&0.93&$\circ$&\textbf{95.56}&$\pm$&0.95&&94.83&$\pm$&0.95&&90.25&$\pm$&0.56&$\bullet$&95.04&$\pm$&0.42&\\
sick&\underline{97.86}&$\pm$&0.33&96.91&$\pm$&0.32&$\bullet$&97.60&$\pm$&0.29&&96.70&$\pm$&0.47&$\bullet$&97.81&$\pm$&0.49&&97.53&$\pm$&0.28&&96.98&$\pm$&0.48&$\bullet$&\textbf{97.93}&$\pm$&0.30&\\
sonar&\textbf{73.55}&$\pm$&6.51&72.26&$\pm$&1.77&&58.71&$\pm$&5.88&$\bullet$&70.97&$\pm$&6.55&&69.68&$\pm$&6.08&&70.32&$\pm$&7.18&&\underline{73.23}&$\pm$&3.34&&72.26&$\pm$&8.10&\\
soybean&\textbf{93.46}&$\pm$&1.32&91.12&$\pm$&1.67&&86.34&$\pm$&2.72&$\bullet$&93.37&$\pm$&1.41&&92.68&$\pm$&2.36&&\textbf{93.46}&$\pm$&0.82&&90.73&$\pm$&1.54&$\bullet$&90.05&$\pm$&1.01&$\bullet$\\
spectrometer&44.40&$\pm$&2.83&\underline{44.53}&$\pm$&3.00&&29.81&$\pm$&4.59&$\bullet$&37.74&$\pm$&4.40&$\bullet$&42.52&$\pm$&3.13&&35.97&$\pm$&3.43&$\bullet$&\textbf{45.41}&$\pm$&2.90&&41.51&$\pm$&2.35&$\bullet$\\
splice&87.40&$\pm$&0.82&95.24&$\pm$&1.01&$\circ$&93.58&$\pm$&2.62&$\circ$&\textbf{95.42}&$\pm$&0.45&$\circ$&\underline{95.36}&$\pm$&0.49&$\circ$&\underline{95.36}&$\pm$&0.49&$\circ$&\underline{95.36}&$\pm$&0.49&$\circ$&85.14&$\pm$&2.18&\\
steel-plates-faults&93.23&$\pm$&3.25&94.40&$\pm$&1.60&&\textbf{99.04}&$\pm$&0.59&$\circ$&91.55&$\pm$&1.41&&88.38&$\pm$&2.59&$\bullet$&90.58&$\pm$&1.39&&\underline{96.22}&$\pm$&0.79&&92.13&$\pm$&2.28&\\
texture&\textbf{96.13}&$\pm$&0.20&78.59&$\pm$&1.32&$\bullet$&94.44&$\pm$&0.54&$\bullet$&95.27&$\pm$&0.21&$\bullet$&\underline{96.11}&$\pm$&0.03&&95.04&$\pm$&0.34&$\bullet$&79.75&$\pm$&1.22&$\bullet$&95.66&$\pm$&0.72&\\
thyroid-disease&\underline{93.64}&$\pm$&0.19&93.02&$\pm$&0.43&$\bullet$&\textbf{93.69}&$\pm$&0.41&&93.32&$\pm$&0.38&$\bullet$&93.35&$\pm$&0.35&$\bullet$&93.35&$\pm$&0.39&$\bullet$&93.31&$\pm$&0.38&&92.75&$\pm$&0.46&$\bullet$\\
vehicle&\textbf{71.65}&$\pm$&1.69&60.71&$\pm$&1.40&$\bullet$&64.17&$\pm$&2.57&$\bullet$&\underline{71.02}&$\pm$&2.08&&69.69&$\pm$&1.24&&70.47&$\pm$&2.06&&59.45&$\pm$&2.88&$\bullet$&67.40&$\pm$&2.43&$\bullet$\\
vote&94.77&$\pm$&1.38&90.15&$\pm$&2.79&$\bullet$&93.85&$\pm$&3.85&&95.08&$\pm$&0.88&&94.62&$\pm$&1.22&&\underline{95.69}&$\pm$&1.77&&90.31&$\pm$&2.70&$\bullet$&\textbf{96.31}&$\pm$&1.26&$\circ$\\
vowel&80.74&$\pm$&3.02&62.96&$\pm$&3.04&$\bullet$&62.83&$\pm$&3.55&$\bullet$&\textbf{92.39}&$\pm$&2.08&$\circ$&81.68&$\pm$&1.08&&\textbf{92.39}&$\pm$&2.55&$\circ$&64.85&$\pm$&2.68&$\bullet$&84.92&$\pm$&1.38&$\circ$\\
waveform-5000&\textbf{81.43}&$\pm$&0.99&79.25&$\pm$&0.75&$\bullet$&\underline{80.95}&$\pm$&1.66&&78.85&$\pm$&0.76&$\bullet$&72.68&$\pm$&0.98&$\bullet$&78.64&$\pm$&1.56&$\bullet$&79.28&$\pm$&0.37&$\bullet$&76.59&$\pm$&0.59&$\bullet$\\
zoo&90.00&$\pm$&3.33&\textbf{95.33}&$\pm$&3.80&$\circ$&52.00&$\pm$&17.1&$\bullet$&\underline{94.00}&$\pm$&1.49&&92.67&$\pm$&2.79&&93.33&$\pm$&2.36&&93.33&$\pm$&2.36&&89.33&$\pm$&5.96&\\
\hline
Ave. Accuracy & \multicolumn{3}{c}{\textbf{83.05}} &  \multicolumn{4}{c}{78.88} &
			\multicolumn{4}{c}{76.53} &
			\multicolumn{4}{c}{82.27} & \multicolumn{4}{c}{80.71} & \multicolumn{4}{c}{81.72} & \multicolumn{4}{c}{79.74} & \multicolumn{4}{c}{81.13}\\
Ave. Rank & \multicolumn{3}{c}{\textbf{2.95}} &  \multicolumn{4}{c}{5.30} &
			\multicolumn{4}{c}{5.29} &
			\multicolumn{4}{c}{3.96} & \multicolumn{4}{c}{4.54} & \multicolumn{4}{c}{4.72} & \multicolumn{4}{c}{4.59} & \multicolumn{4}{c}{4.65}\\
W/T/L & \multicolumn{3}{c}{-} &  \multicolumn{4}{c}{30/26/4} &
			\multicolumn{4}{c}{29/23/8} &
			\multicolumn{4}{c}{20/34/6} & \multicolumn{4}{c}{25/27/8} & \multicolumn{4}{c}{23/31/6} & \multicolumn{4}{c}{30/24/6} & \multicolumn{4}{c}{14/38/8}\\
$p$-value & \multicolumn{3}{c}{-} &  \multicolumn{4}{c}{\textbf{<0.001}} &
			\multicolumn{4}{c}{\textbf{<0.001}} &
			\multicolumn{4}{c}{\textbf{0.003}} & \multicolumn{4}{c}{\textbf{0.003}} & \multicolumn{4}{c}{\textbf{0.001}} & \multicolumn{4}{c}{\textbf{0.001}} & \multicolumn{4}{c}{\textbf{<0.001}}\\\hline
\end{tabular*} \scriptsize \par}
\end{table}

\subsection{Performance Comparison with Baselines}
\label{sec: Experimental Results and Analysis}
We randomly split each of the UCI datasets into training and test sets according to the ratio of 7:3 for five times. The averaged classification accuracies on test sets are reported as the evaluation metric. Table \ref{tab:AccuracyComparison} compares the classification accuracies of NeuralKDB and baseline classifiers on the 60 UCI datasets. For each dataset, the best and second best performers are respectively highlighted by \textbf{boldface} and \underline{underline}. For rigorous comparison, we conduct two-tailed paired t-tests between NeuralKDB and baselines on each dataset and use $\bullet$ ($\circ$) to denote that NeuralKDB is significantly superior (inferior) to the compared baselines at the $0.05$ significance level. We summarize the performance comparison at the bottom of Table \ref{tab:AccuracyComparison}. ``Ave. Accuracy'' gives the averaged accuracies of NeuralKDB and baseline classifiers over all 60 datasets. ``Ave. Rank'' provides the averaged ranks of NeuralKDB and baselines over 60 datasets, obtained by the Friedman test \citep{Demsar06}. ``W/T/L'' lists the number of times NeuralKDB wins, ties and loses when comparing with the corresponding baselines across all datasets. In addition, we also conduct the Wilcoxon signed-ranks tests \citep{Demsar06} to statistically compare the overall performance of NeuralKDB and baseline classifiers. The resulting $p$-values are reported at the last row of Table \ref{tab:AccuracyComparison}, where a $p$-value smaller than $0.05$ implies that NeuralKDB is significantly better than the corresponding baseline. 

As is shown by Table \ref{tab:AccuracyComparison}, it is evident that NeuralKDB consistently outperforms all the competing methods with the largest average accuracy score and smallest average rank. The better performance of NeuralKDB over the conventional BNCs (KDF, KDB, TAN and NB) are attributed to the superior capability of the distributional representations learned by NeuralKDB in modeling the high-order feature dependencies, which is transferable to extrapolating the occurrence probabilities of unseen test samples. The comparison with NeuralNB and NN further necessitates distributional representation learning. Though NeuralNB and NN learn feature value representations by capturing the feature-label correlations, the lack of feature-feature dependence semantics restricts their performance. It is noteworthy that NeuralKDB outperforms the competitive classifier RF. The good performance of RF is mainly contributed by the ensemble operation, while NeuralKDB is a single model. We can expect that the performance of NeuralKDB would be improved further, if we reinforce NeuralKDB with ensemble learning.

\begin{figure*}[t]
  \centering
  \begin{subfigure}[b]{0.24\linewidth}
    \centering
    \includegraphics[width=\linewidth]{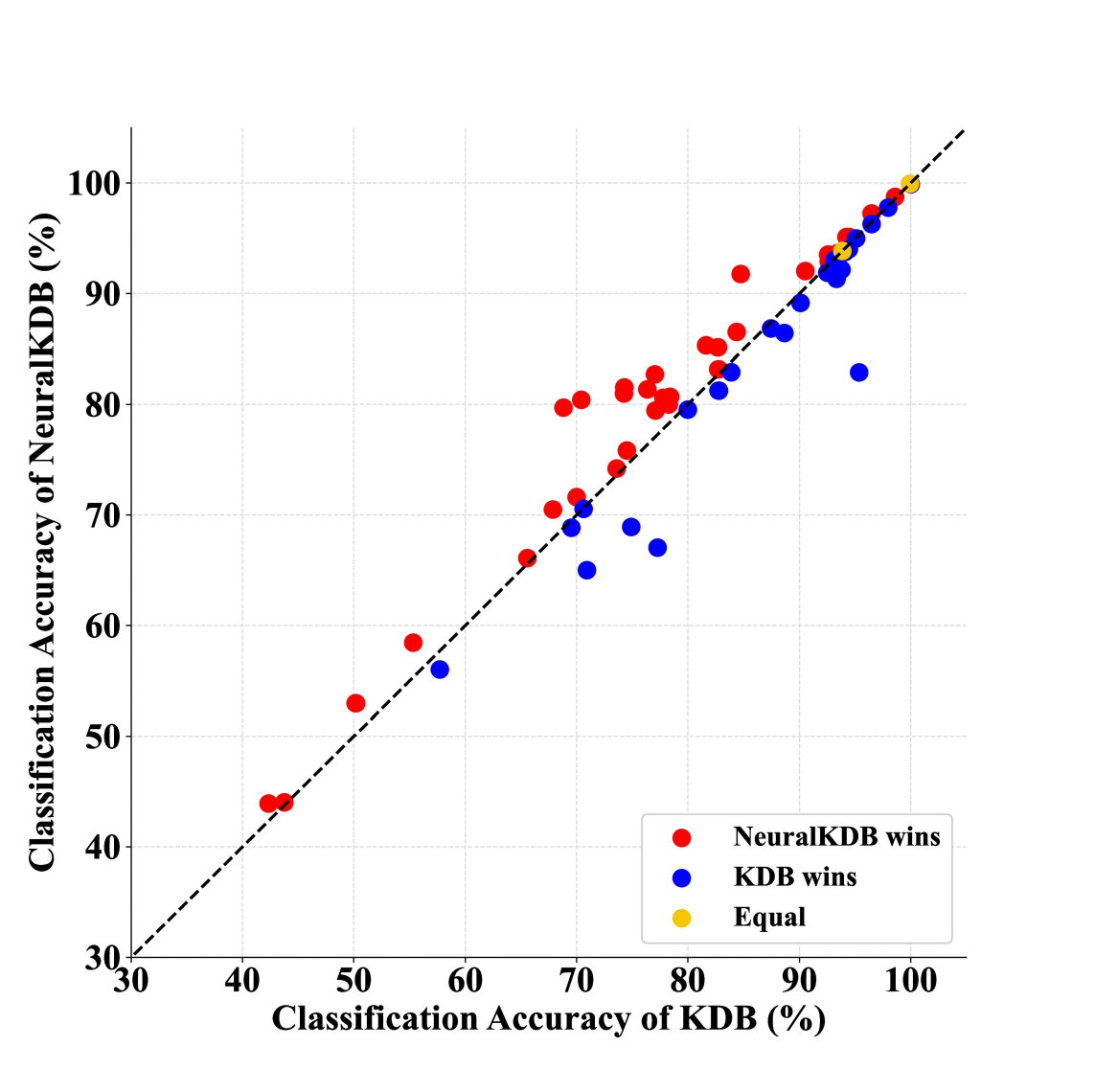}
    \caption{$k=1$}
    \label{fig:K1}
  \end{subfigure}
  \begin{subfigure}[b]{0.24\linewidth}
    \centering
    \includegraphics[width=\linewidth]{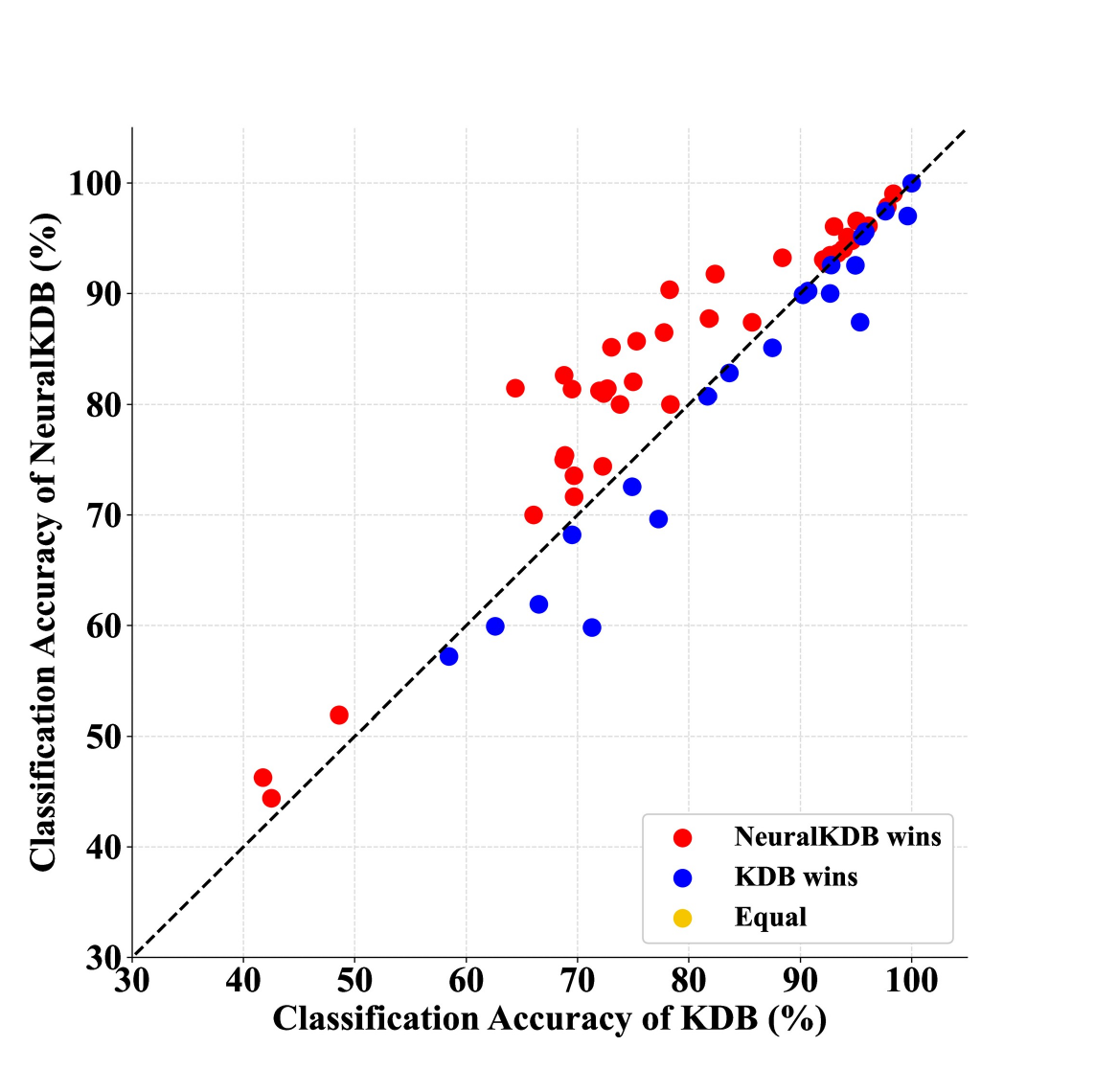}
    \caption{$k=2$}
    \label{fig:K2}
  \end{subfigure}
  \begin{subfigure}[b]{0.24\linewidth}
    \centering
    \includegraphics[width=\linewidth]{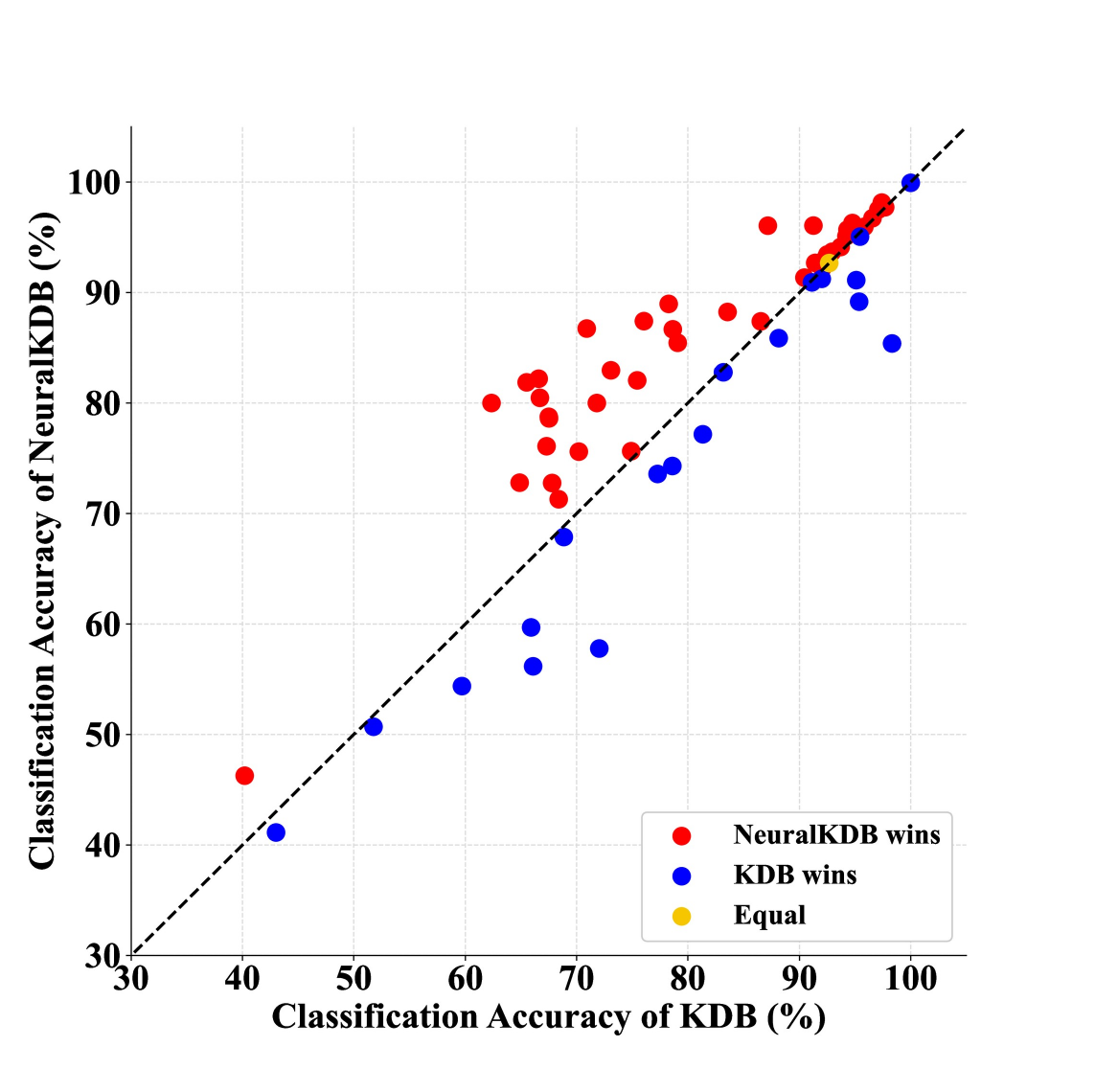}
    \caption{$k=3$}
    \label{fig:K3}
  \end{subfigure}
  \begin{subfigure}[b]{0.24\linewidth}
    \centering
    \includegraphics[width=\linewidth]{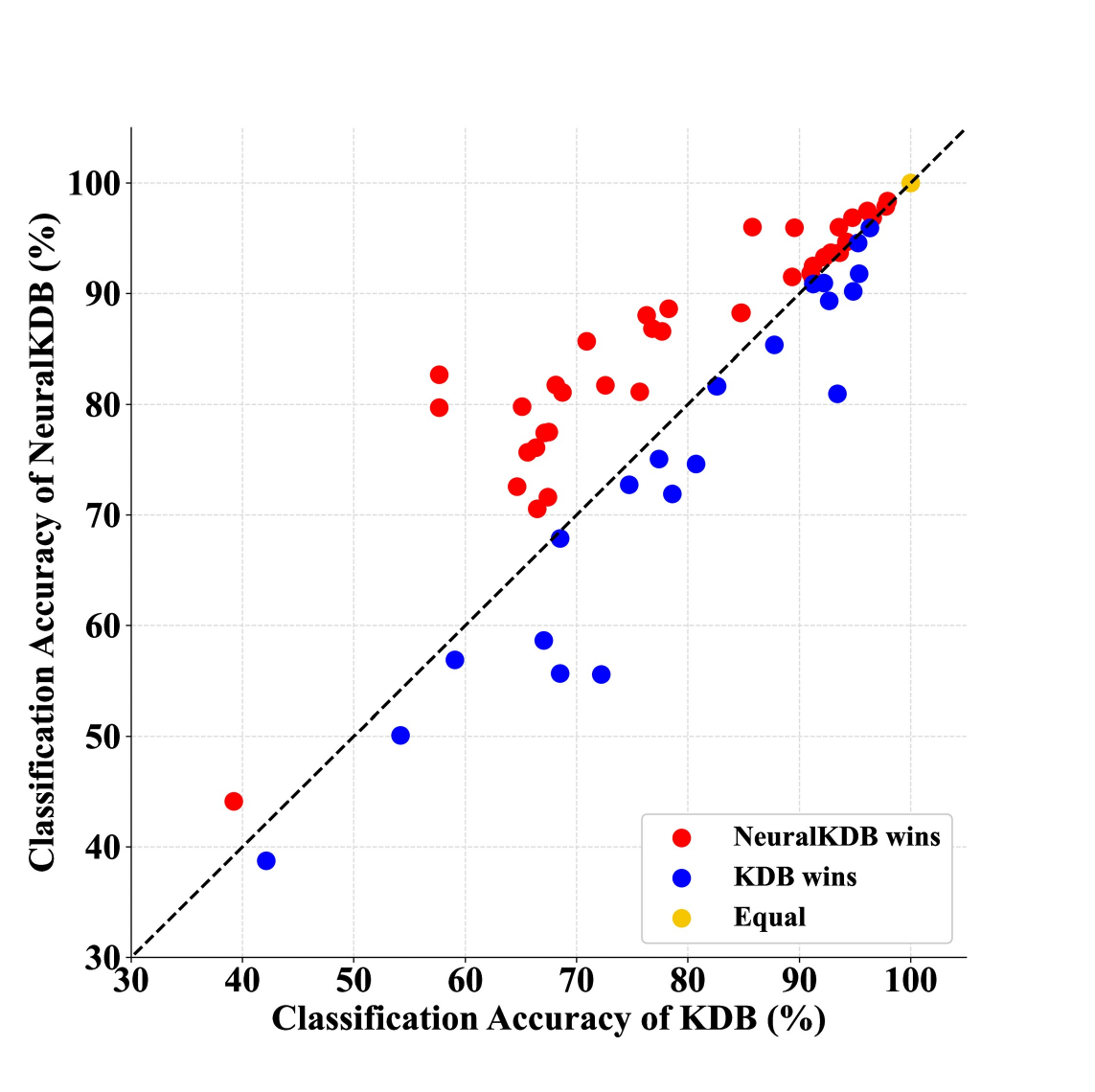}
    \caption{$k=4$}
    \label{fig:K4}
  \end{subfigure}
  \caption{Scatter plots for the performance comparison between NeuralKDB and KDB with different $k$ values.}
  \label{fig:ACC}
\end{figure*}

\begin{table}[!t]
\centering
\addtolength{\tabcolsep}{+2.5pt}
\caption{Overall performance comparison between NeuralKDB and KDB with different $k$ values.}	 
     \begin{tabular*}{0.75\textwidth}{ccccccc}
        \hline 
          &\multicolumn{2}{c}{Ave. Accuracy}&\multicolumn{2}{c}{Ave. Rank}&\multirow{2}{*}{W/T/L}&\multirow{2}{*}{$p$-value} \\ \cmidrule(r){2-3} \cmidrule(r){4-5}
          &NeuralKDB&KDB&NeuralKDB&KDB&& \\\hline
          $k=1$&\textbf{82.42}&81.72&\textbf{1.43}&1.57&12/39/9&\textbf{0.043}\\
          $k=2$&\textbf{83.05}&80.71&\textbf{1.35}&1.65&25/27/8&\textbf{0.003}\\
          $k=3$&\textbf{82.68}&80.18&\textbf{1.32}&1.68&25/26/9&\textbf{0.005}\\
          $k=4$&\textbf{82.21}&79.53&\textbf{1.36}&1.64&24/23/13&\textbf{0.014}\\
         \hline
        \end{tabular*}
        \label{tab:summary2}
\end{table}

\begin{figure}[!t]
  \centering
  \begin{subfigure}[b]{0.35\linewidth}
    \centering
    \includegraphics[width=\linewidth]{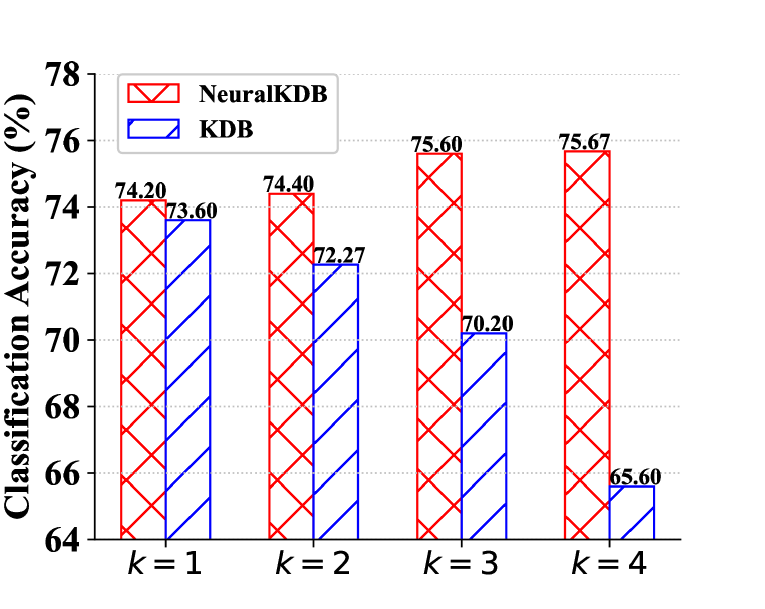}
    \caption{credit-g}
    \label{fig:batch}
  \end{subfigure}
  \begin{subfigure}[b]{0.35\linewidth}
    \centering
    \includegraphics[width=\linewidth]{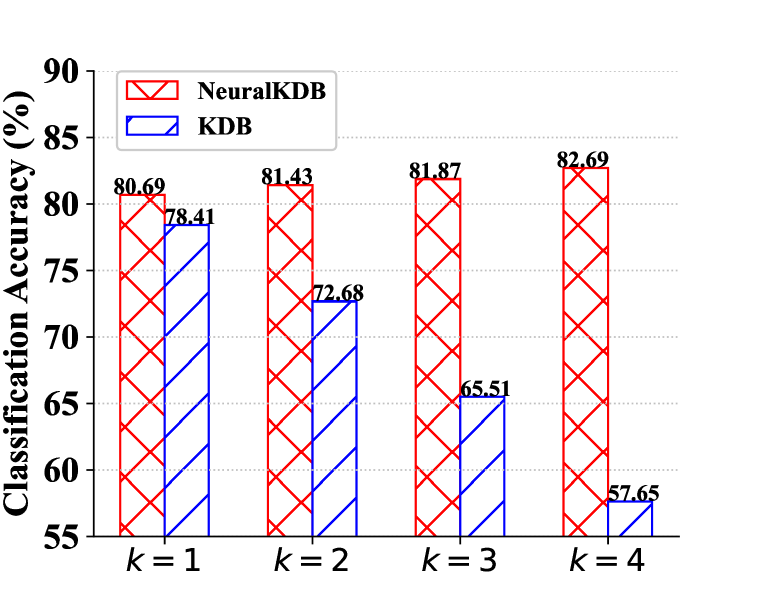}
    \caption{waveform-5000}
    \label{fig:embedding}
  \end{subfigure}
  \caption{Performance comparison between NeuralKDB and KDB on the two large datasets with different $k$ values.}
  \label{fig:Case}
\end{figure}

\begin{figure*}[t]
  \centering
  \begin{subfigure}[b]{0.3\linewidth}
    \centering
    \includegraphics[width=\linewidth]{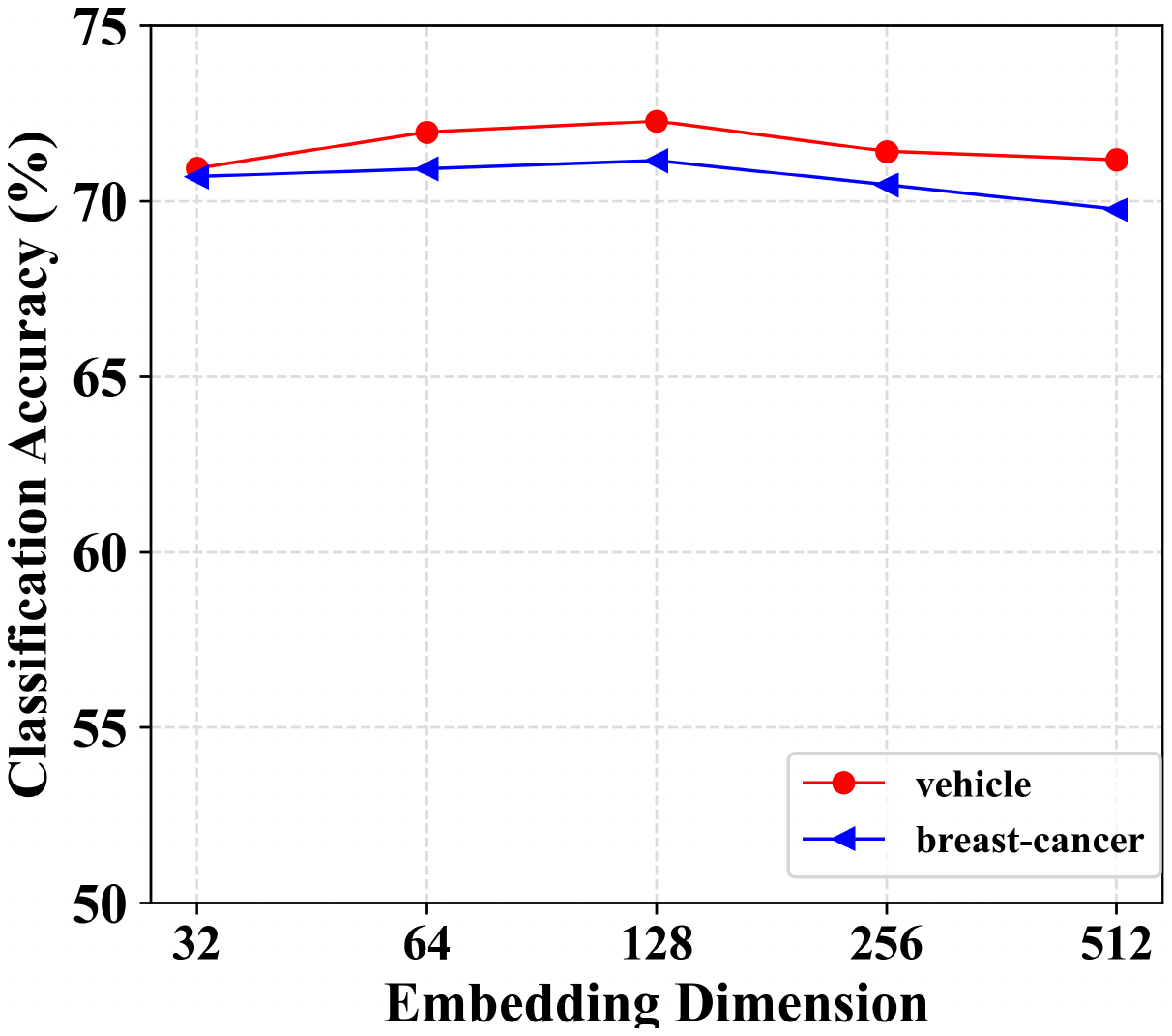}
    \caption{Embedding Dimension}
    \label{fig:batch}
  \end{subfigure}
  \begin{subfigure}[b]{0.3\linewidth}
    \centering
    \includegraphics[width=\linewidth]{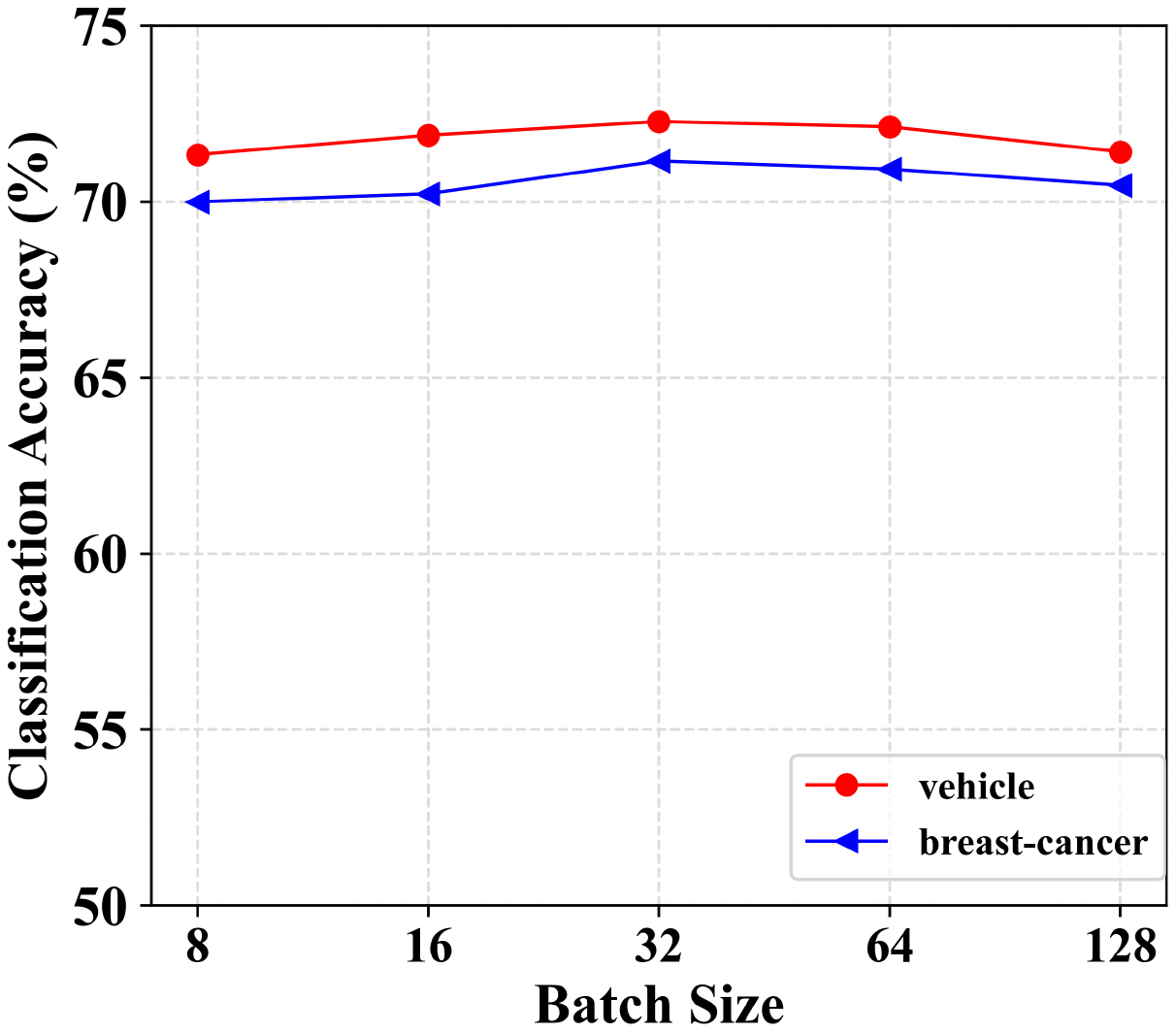}
    \caption{Batch Size}
    \label{fig:embedding}
  \end{subfigure}
  \begin{subfigure}[b]{0.3\linewidth}
    \centering
    \includegraphics[width=\linewidth]{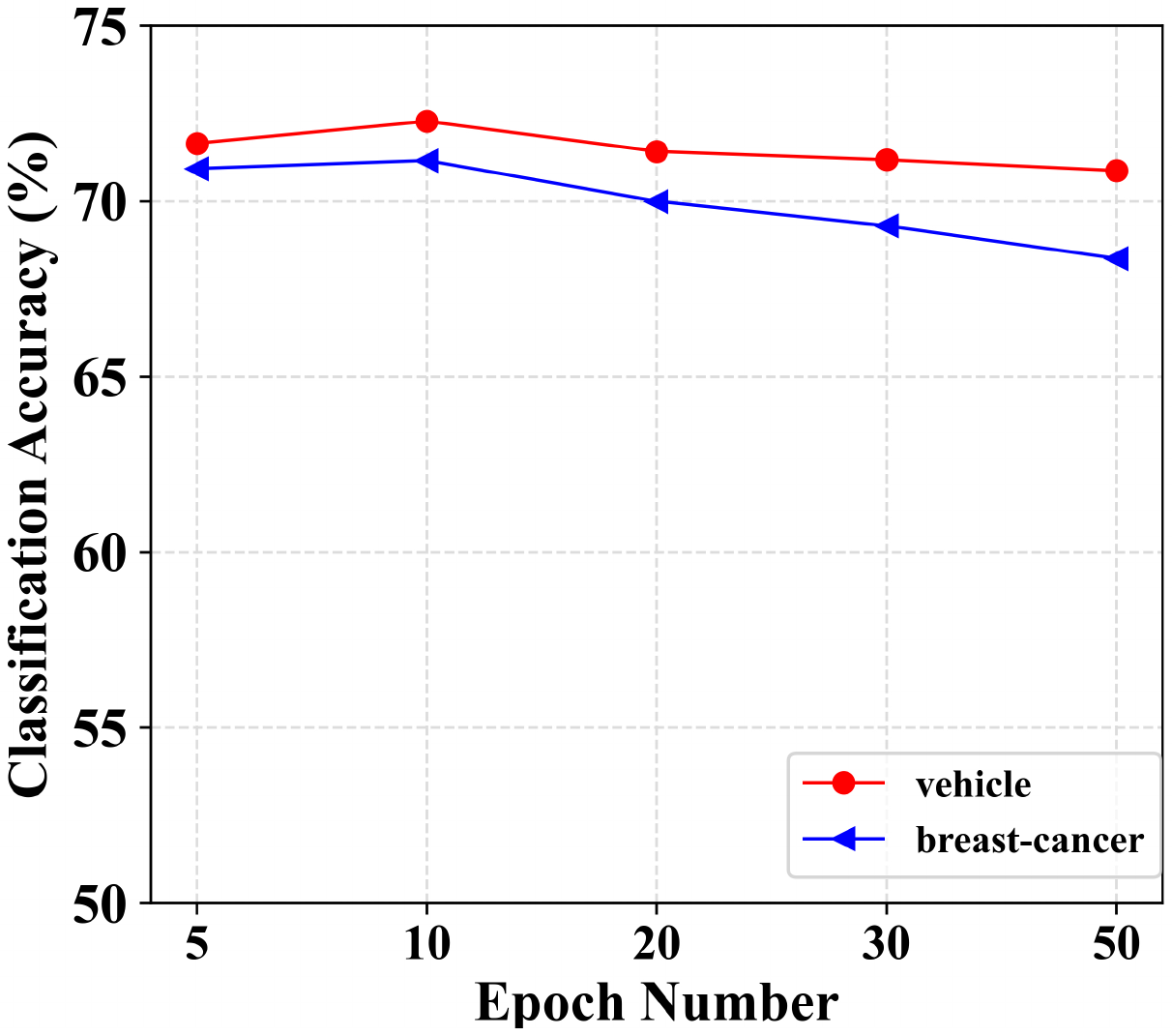}
    \caption{Epoch Number}
    \label{fig:epoch}
  \end{subfigure}
  \caption{The sensitivity of NeuralKDB with regard to the three hyperparameters.}
  \label{fig:para}
\end{figure*}

\subsection{Performance with Varying Dependencies}
\label{sec: Discussion}
To test the robustness of NeuralKDB against data sparsity, we perform a case study to compare the performance of NeuralKDB and KDB at different levels of dependencies by varying the dependence order $k$ from 1 to 4. Figure \ref{fig:ACC} shows the scatter plots of the accuracies of NeuralKDB and KDB on the 60 UCI datasets, where the Y-axis and X-axis respectively represent the accuracies of NeuralKDB and KDB. In the plots, the red, blue and yellow dot points indicate the datasets on which NeuralKDB achieves a higher, lower or equal classification accuracy in comparison with KDB. Table \ref{tab:summary2} summarizes more details about the comparison, including the ``W/T/L'' counts of NeuralKDB in comparison with KDB, and $p$-values of the Wilcoxon tests \citep{Demsar06}. 

From Figure \ref{fig:ACC} and Table \ref{tab:summary2}, we can find that NeuralKDB consistently performs better than KDB with all dependence orders. When the dependence order $k$ increases from 1 to 2, both NeuralKDB and KDB go through a performance gain, as the benefit of 2-order dependence modeling. However, we can observe a performance decrease trend for both methods when $k$ increases from 2 to 4. This is caused by the fact that many of the 60 datasets are in small scales, with a small number of samples and low feature dimensions, making 2-order dependency better fit their characteristics. Despite this, NeuralKDB decreases by a smaller margin than KDB, which proves NeuralKDB's robustness against the data sparsity issue.

To better evaluate NeuralKDB's ability in capturing high-order feature dependencies, we investigate the classification performance on two large datasets, ``credit-g'' and ``waveform-5000''. Figure \ref{fig:Case} compares the performance of NeuralKDB and KDB on the two datasets. From Figure \ref{fig:Case}, we can observe that NeuralKDB performs increasingly better with the increase of the dependence order, while KDB goes through a dramatic performance drop. The investigation verifies that NeuralKDB excels in capturing high-order feature dependencies that exist in large datasets.

\subsection{Parameter Sensitivity Study}
\label{sec: Sensitivity Analysis}
In this section, we study NeuralKDB's sensitivity regarding to its three hyperparameters: the embedding dimension $d$, as well as the batch size and epoch number used for model training. We take turns to select each of the three hyperparameters, make it vary in a proper range while fix the remaining two hyperparameters as default values, and study the performance change of NeuralKDB on the two datasets, ``vehicle'' and ``breast-cancer''. Figure \ref{fig:para} plots the performance change with varying hyperparameters. From Figure \ref{fig:para}, we can find that the performance of NeuralKDB remains relatively stable with the change of the hyperparameters.

\section{Conclusion and Future Work}
In this paper, we propose a novel paradigm to design the next-generation BNCs, by using distributional representation learning to overcome the parameter explosion and data sparsity issues, given its success in natural language processing and graph learning. The feasibility and effectiveness of the proposed paradigm has been verified with the invented NeuralKDB classifier, which excels in capturing high-order feature dependencies and achieves significant better classification performance than traditional BNCs. We believe that distributional representation learning could be more powerful in revolutionizing the design of BNCs, by using multi-layer neural networks to learn deep feature value representations, which would better capture the non-linear feature correlations in real-world data.  

\bibliography{NeuralKDB}
\end{document}